\documentclass{article} 
\usepackage{iclr2027_conference,times}

\usepackage{amsmath,amsfonts,bm}

\def\eqref#1{equation~\ref{#1}}

\def\1{\bm{1}}

\DeclareMathAlphabet{\mathsfit}{\encodingdefault}{\sfdefault}{m}{sl}
\SetMathAlphabet{\mathsfit}{bold}{\encodingdefault}{\sfdefault}{bx}{n}

\usepackage{url}
\usepackage{booktabs}
\usepackage{multirow}
\usepackage{graphicx}
\usepackage{bbm}
\usepackage{subcaption}
\usepackage{hyperref}
\usepackage{algorithm}
\usepackage{algpseudocode}
\usepackage{amsmath,amssymb}
\usepackage{wrapfig}
\usepackage{xcolor}
\newcommand{\rev}[1]{#1}

\title{S$^2$LMR: Supervised Sequential Long-Context Multi-Step Retriever}

\author{Qiang Hu \& Kaiyi Ji \\
Department of Computer Science and Engineering \\
University at Buffalo \\
Buffalo, NY, USA \\
\texttt{\{qhu5,kaiyiji\}@buffalo.edu}\\[8pt]
{\normalfont\small\textbf{Project page:} \url{https://fujump.github.io/s2lmr_page/}}
}

\iclrfinalcopy 

\begin{document}

\maketitle
\fancyhead{}
\lhead{Preprint}

\begin{abstract}
Multi-step retrieval over long contexts is a key capability for retrieval-augmented generation, but strong existing approaches are costly and complex to train.
Supervised retrievers are far simpler and cheaper, yet they fall well behind in this setting.
We show that this gap stems not from supervised training itself, but from three misalignments between training and inference: easy training negatives, noisy retrieval prefixes, and position-agnostic scoring.
Guided by this analysis, we introduce S$^2$LMR, a simple yet effective supervised sequential retriever that models multi-step retrieval as the evolution of an explicit textual state encoded by a single shared Transformer.
On long-context reasoning, multi-hop QA, and needle-in-a-haystack benchmarks with context lengths ranging from 4K to 10M tokens, S\(^2\)LMR achieves performance comparable to state-of-the-art methods, substantially outperforming prior supervised retrievers. Notably, it requires only \(\sim\)2 hours of training on a single A6000 for contexts of up to 10M tokens, while remaining robust across a wide range of hyperparameter settings.
\end{abstract}

\section{introduction}
Retrieval-Augmented Generation (RAG) has become a standard paradigm for augmenting large language models (LLMs) with external knowledge, improving factuality without requiring large knowledge sources to be directly included in the model context~\citep{lewis2020retrieval,guu2020retrieval,izacard2021leveraging,gao2023retrieval}. Most existing RAG studies focus on single-step retrieval, which is often sufficient for relatively simple tasks such as Needle-in-a-Haystack~\citep{hsieh2024ruler}. 
However, single-step retrieval is insufficient for emerging scenarios that require multiple rounds of interaction with the context, such as multi-hop question answering~\citep{yang2018hotpotqa,trivedi2022musique} and reasoning over millions of tokens~\citep{kuratov2024babilong}. 
These settings require the retriever to progressively gather relevant evidence across multiple steps. 

There are several types of approaches for multi-step retrieval. 
Graph-based approaches build structured graphs over retrieved information to connect evidence across documents, but often incur additional overhead from graph construction and traversal~\citep{li2024graphreader,mao2021generation}. Some works use LLM agents to guide multi-step retrieval through generated reasoning or instructions~\citep{anokhin2024arigraph,singh2025agentic}. However, repeated LLM calls increase inference cost, while noisy evidence can misguide later retrieval steps.  
Another line of work fine-tunes LLMs to interact with retrieval tools~\citep{jin2025search,song2025r1}, but fine-tuning LLMs can be expensive and difficult to scale. 

More recently, \citet{sorokin2026q} proposed a more resource-efficient approach, Q-RAG, that uses reinforcement learning (RL) to fine-tune the embedder model, achieving state-of-the-art results on multiple long-context multi-step benchmarks with contexts of up to 10 million tokens. However, compared with conventional supervised retrieval using bi-encoders or cross-encoders~\citep{xiong2020answering,zhang2024end}, this approach introduces additional RL components, such as value and target networks, as well as temporal-difference updates, increasing implementation complexity and introducing additional hyperparameters. This raises an intriguing but non-trivial question: 
\begin{center}
   {\em Can we achieve effective multi-step retrieval over million-token contexts using only a simple supervised learning framework?} 
\end{center}
Through extensive analysis, we identify three misalignments that explain why supervised retrievers fall far behind recent methods such as Q-RAG:
(i)~training negatives are easy, unrelated passages, whereas test-time distractors come from the same context as the gold passage;
(ii)~later hops are trained on the model's own, sometimes wrong, retrieval prefixes, which injects noise into their supervision;
and (iii)~the scorer ignores where a candidate lies relative to the passages already retrieved, even though this position strongly predicts the next relevant passage.

Motivated by these observations, we propose the \textbf{S}upervised \textbf{S}equential \textbf{L}ong-Context \textbf{M}ulti-Step \textbf{R}etriever (S$^2$LMR), a supervised bi-encoder designed to align training with long-context multi-step retrieval.
S$^2$LMR addresses the identified mismatches through same-context negatives, gold-prefix next-evidence supervision, and history-adaptive relative positions, using a single shared Transformer encoder with an explicit \emph{textual state} that evolves across retrieval hops.
Across (ultra)-long context benchmarks and multi-hop QA, S$^2$LMR matches or outperforms the state-of-the-art RL-based retriever Q-RAG~\citep{sorokin2026q}. 
Our main contributions are as follows: \begin{itemize} 
\item We propose S$^2$LMR, a supervised sequential retriever built on a single shared Transformer encoder and a simple contrastive supervision. 
\item On long-context commonsense reasoning, multi-hop QA, and needle-in-a-haystack tasks with contexts from 4K to 10M tokens, S$^2$LMR matches or outperforms the state-of-the-art Q-RAG.
It also substantially outperforms Beam Retriever, the strongest existing supervised retriever.
On BabiLong QA3, S$^2$LMR keeps fact EM above 0.95 at up to 10M tokens, while Beam Retriever reaches only 0.2--0.3.

\item \rev{We further show that S$^2$LMR offers substantial advantages in training efficiency and stability.
For example, on BabiLong QA3, S$^2$LMR reaches its peak performance in 1.4 GPU hours and completes training in 2.25h on a single A6000,
compared with 3h and 54.4h for Q-RAG}, and more than 400 GPU hours for Beam Retriever.
Moreover, compared with Q-RAG, S$^2$LMR exhibits more stable training dynamics and is substantially more robust to a wide range of hyperparameter choices, including learning rate and batch size.
\end{itemize}

\section{Related Works}
\label{sec:related}

\textbf{\em Multi-step retrieval.}
Multi-hop retrieval methods~\citep{das2019multi,qi2019answering} extend single-step dense retrieval~\citep{karpukhin2020dense,izacard2021unsupervised} by conditioning each retrieval decision on previously retrieved evidence. MDR~\citep{xiong2020answering} uses a bi-encoder and iteratively augments the query with retrieved passages, while BeamRetriever~\citep{zhang2024end} combines a cross-encoder with beam search for stronger multi-hop retrieval at higher inference cost. Q-RAG~\citep{sorokin2026q} further studies embedder-based multi-step retrieval in long-context settings. Another line of work uses LLM agents to interleave retrieval with generated reasoning~\citep{yao2022react,trivedi2023interleaving,shao2023enhancing,asai2024self,singh2025agentic,jin2025search,song2025r1}. These methods provide flexible reasoning but require repeated LLM inference across retrieval steps. S$^2$LMR focuses instead on more efficient sequential retrieval with a lightweight shared encoder.

\textbf{\em Retrieval for long-context tasks.}
Long-context benchmarks such as RULER~\citep{hsieh2024ruler} and BabiLong~\citep{kuratov2024babilong} evaluate reasoning over contexts ranging from thousands to millions of tokens. Existing approaches improve long-context processing through architectural changes, including recurrent memory~\citep{bulatov2022recurrent,behrouz2026titans,behrouz2025atlas,rodkin2024associative}, state-space models~\citep{gu2023mamba}, and positional encoding adaptation~\citep{shang2025longrope2}. An alternative is to first retrieve a small subset of relevant content from the long context~\citep{jiang2024longllmlingua,jiang2024longrag}. Q-RAG~\citep{sorokin2026q} follows this direction and shows that learned chunk embeddings can support multi-step retrieval over very long contexts. S$^2$LMR builds on this retrieval-based paradigm while introducing history-adaptive retrieval states and candidate positions for simple supervised multi-step retrieval.

\textbf{\em Supervised training for dense retrieval.}
Dense retrievers are commonly trained with contrastive objectives~\citep{oord2018representation,karpukhin2020dense}. DPR~\citep{karpukhin2020dense} trains a bi-encoder using in-batch negatives together with hard negatives from term-based retrieval, and multi-hop methods such as MDR~\citep{xiong2020answering} largely follow the same training strategy. Such objectives were originally developed for open-domain retrieval, where negatives often come from other queries or documents. In long-context multi-step retrieval, however, the main competitors at inference are chunks from the same context, and the correct target can also depend on previously retrieved evidence. Q-RAG~\citep{sorokin2026q} instead formulates retrieval as a Markov decision process and trains with temporal-difference learning~\citep{mnih2015human}. In contrast, S$^2$LMR retains simple supervised contrastive training while aligning the candidate set, retrieval state, and relative positions with those encountered during inference.

\section{Methods}
\label{sec:method}

\begin{figure}
    \centering
    \includegraphics[width=1\linewidth]{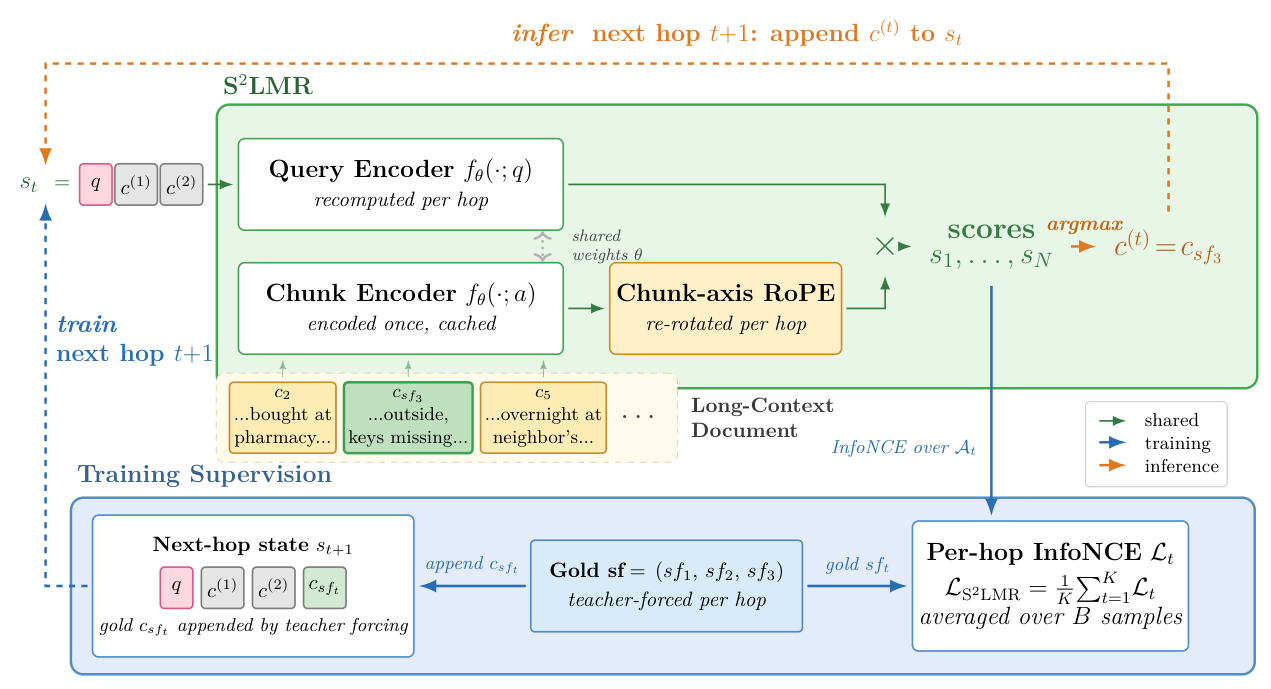}
    \caption{Overview of S$^2$LMR, which performs sequential multi-hop retrieval with a shared bi-encoder. The initial state $s_0$ contains the query $q$, and all chunks of the long context are embedded and cached at the start. At each hop $t$, S$^2$LMR embeds the current state $s_t$, re-rotates cached chunk embeddings according to the retrieval history, and scores candidate chunks by their inner products with the state embedding. During training (blue), gold chunks provide per-hop supervision; during inference (orange), S$^2$LMR greedily selects and appends the highest-scoring chunk. 
    }
    \label{fig:overview}
\end{figure}

\subsection{Revisiting Supervised RAG for Long-Context Multi-Step Retrieval}

Supervised retrievers remain attractive for multi-step retrieval because they are efficient and can be trained directly from retrieval labels. Two representative approaches are MDR~\citep{xiong2020answering} and Beam Retriever (BR)~\citep{zhang2024end}. MDR performs sequential dense retrieval with a bi-encoder, while BR scores each new candidate together with the question and previously retrieved passages. These methods were mainly developed for open-domain or relatively small candidate sets. Long-context retrieval is different: the model must search over many passages from the same context, where distractors can be highly similar to the correct evidence.

To test how well existing designs transfer to this setting, we first train a bi-encoder with the standard DPR/MDR recipe on the challenging long-context BabiLong QA3 benchmark~\citep{kuratov2024babilong}. It achieves only 1.6\% fact exact match (fact-EM), substantially below recent multi-step retrievers. We identify three issues that become especially important as the context and candidate pool grow.

\label{sec:motivation_ssr}

\begin{figure}[t]
    \centering
    \begin{subfigure}[t]{0.325\textwidth}
        \includegraphics[width=\linewidth]{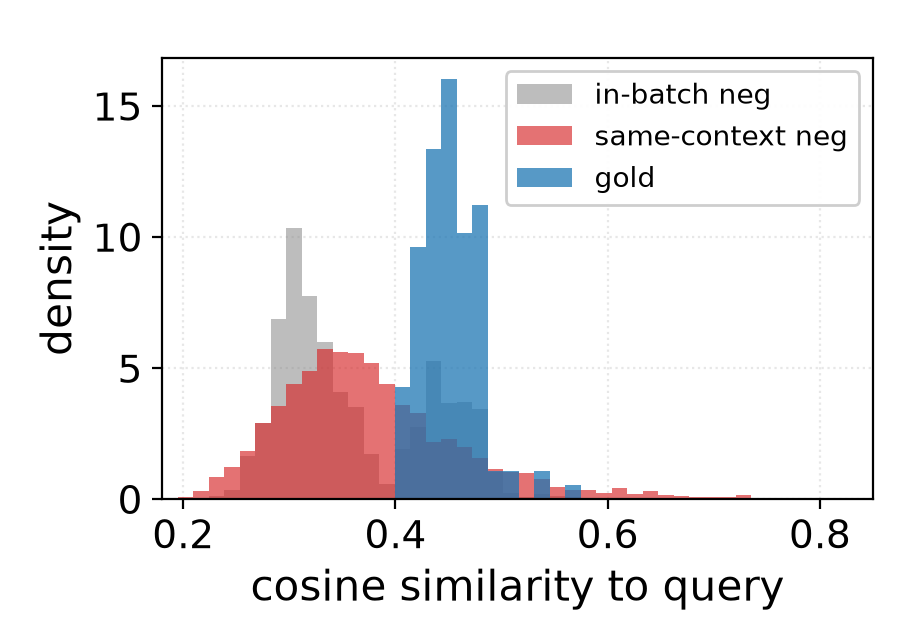}
        \caption{}
        \label{fig:m1_evidence}
    \end{subfigure}
    \hfill
    \begin{subfigure}[t]{0.325\textwidth}
        \includegraphics[width=\linewidth]{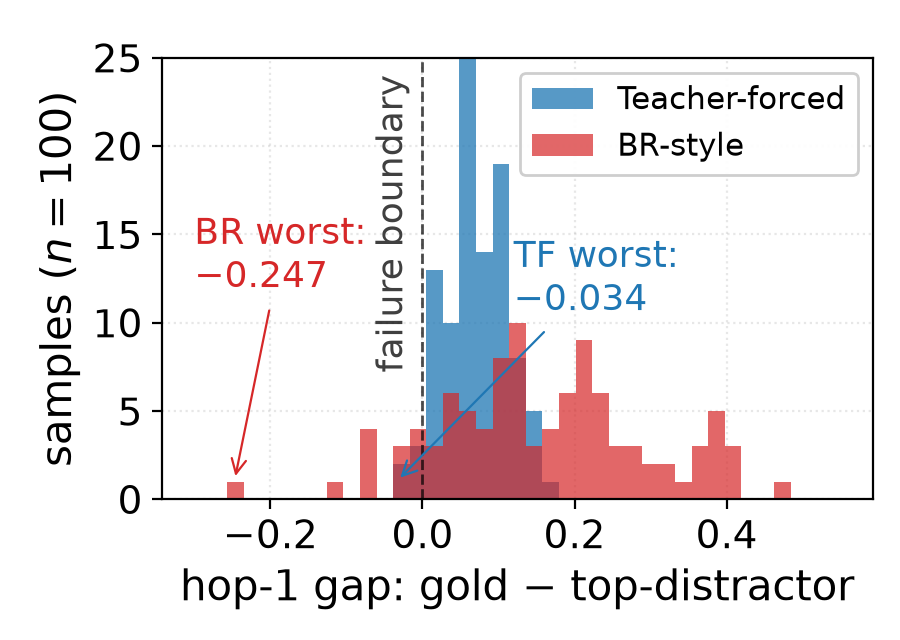}
        \caption{}
        \label{fig:m2_margin}
    \end{subfigure}
    \hfill
    \begin{subfigure}[t]{0.325\textwidth}
        \includegraphics[width=\linewidth]{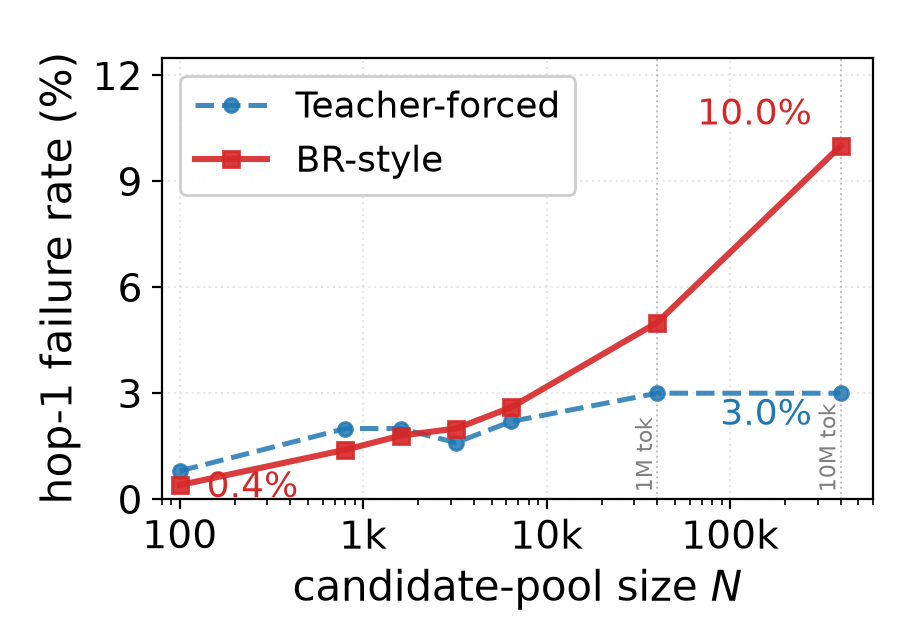}
        \caption{}
        \label{fig:m2_failure}
    \end{subfigure}
    \caption{
    Evidence for Issues 1 and 2.
    (\subref{fig:m1_evidence})~Hop-1 cosine distributions under MDR-style training: gold vs.\ same-context and in-batch negatives.
    (\subref{fig:m2_margin})~Distribution of the hop-1 gold-vs-distractor cosine gap at ns${=}6.4$k.
    (\subref{fig:m2_failure})~Hop-1 failure rate as the candidate pool grows to ${\sim}$10M tokens. 
    }
    \label{fig:motivation}
\end{figure}

\textbf{Issue 1: Training negatives do not match hard same-context distractors.}
MDR mainly learns from negatives associated with other queries, which are often easier than distractors from the same context. In long-context retrieval, however, incorrect passages may mention the same entities, describe related events, and use very similar language. BR uses candidates from the same example, but its original training setting contains only 10--20 candidates, far fewer than the thousands of chunks that may appear in long contexts. Existing training therefore provides limited supervision for distinguishing the gold passage from many highly similar distractors. Figure~\ref{fig:m1_evidence} illustrates this problem: on our 128 evaluation examples, an MDR-style model ranks at least one same-context distractor above the correct first-hop passage in every case.

\textbf{Issue 2: Predicted prefixes make multi-hop supervision noisy.}
BR trains later hops using passages selected by earlier hops. If an earlier prediction is wrong, later hops are  trained on an incorrect retrieval history. This makes the supervision noisier and can reduce the score gap between the gold passage and strong distractors. On 100 held-out QA3 examples with 400k candidates each ($\sim$10M tokens), BR-style training gives a {\bf negative} gold-vs-distractor margin on $12\%$ of examples, with a worst margin of $-0.247$, while a {\em matched model trained} with ground-truth prefixes flips on only $4\%$ of examples, with a worst value of $-0.034$ (Fig.~\ref{fig:m2_margin}). The problem becomes more serious as the candidate pool grows: from 100 to 400k passages ($\sim$10M tokens), the hop-1 failure rate rises from $0.4\%$ to $10\%$ and end-to-end EM drops from $0.994$ to $0.880$, while the matched control remains below $3\%$ (Fig.~\ref{fig:m2_failure}).


\begin{table}[t]
\centering
\small
\setlength{\tabcolsep}{8pt}
\caption{
Per-hop retrieval failure of a fine-tuned chunk-position-agnostic bi-encoder on BabiLong QA3 at $N{=}100$.
$\text{margin}_t =
\cos(q_t,\text{gold})_{\max}
-
\cos(q_t,\text{distractor})_{\max}$.
See Appendix~\ref{app:pe_bottleneck}.
}
\begin{tabular}{lcccc}
\toprule
Metric & hop-1 & hop-2 & hop-3 & K=3 \\
\midrule
$P(\text{margin} \le 0)$ & 0.66 & 0.53 & 0.80 & --- \\
Per-hop pick success & 0.34 & 0.47 & 0.20 & \textbf{8.5\%} \\
\bottomrule
\end{tabular}
\label{tab:pe_bottleneck}
\end{table}

\textbf{Issue 3: Retrieval scores ignore the chunk position.}
Neither MDR nor BR explicitly uses where a candidate appears in the  document. This loses useful information in long-context tasks, where relevant facts are often positionally related. On BabiLong QA3, the next relevant fact lies within 25\% of the previous pick's position in 90\% of examples, with a median offset of only 5\%. For example, after retrieving a fact at chunk 48 for the question ``Where was the apple before the bathroom?'', the word ``before'' suggests that the next relevant fact should appear earlier in the document. A position-agnostic retriever cannot directly use this signal. Table~\ref{tab:pe_bottleneck} shows the consequence: even after fine-tuning, the model often assigns similar scores to the gold passage and its strongest distractor, and per-hop success remains only 20\%--47\%, \rev{limiting three-hop retrieval EM to 8.5\%.}

\subsection{Supervised Sequential Long-Context Multi-Step Retriever}
\label{sec:ssr}
Motivated by Sec.~\ref{sec:motivation_ssr}, we propose S$^2$LMR, a supervised sequential retriever for long-context multi-step retrieval. It aligns training with inference along three complementary dimensions: the candidate set used for contrastive supervision, the sequential retrieval state conditioned on previous evidence, and the history-adaptive relative positions used to score candidates.

S$^2$LMR uses a shared encoder $f_\theta:\mathcal{X}\rightarrow\mathbb{R}^d$ to represent both retrieval states and candidate chunks. Given a text input $x$, we mean-pool its final-layer hidden states over non-padding tokens and apply $\ell_2$ normalization to obtain $f_\theta(x)$. Each example consists of a query $q$ and a context divided into $N$ chunks $c_1,\ldots,c_N$, indexed by $\mathcal{C}=\{1,\ldots,N\}$.

\textbf{Sequential retrieval state.}
S$^2$LMR retrieves one chunk per hop, and each hop is conditioned on the chunks
retrieved before it. Let $c^{(t)}=c_{i_t}$ be the chunk retrieved at hop $t$,
where $i_t$ is its document index, and let $\pi$ be the permutation of
$\{1,\ldots,t-1\}$ with $i_{\pi_1}<\cdots<i_{\pi_{t-1}}$. The state at hop $t$
is
\begin{align}
s_t = \big[\, q,\; c^{(\pi_1)},\; \ldots,\; c^{(\pi_{t-1})} \,\big],
\label{eq:state}
\end{align}
with $s_1=q$, where $[\cdot]$ denotes concatenation with a fixed separator
token such as \texttt{[SEP]}. The permutation only determines how the
retrieved chunks are arranged in the input. Chunks are still retrieved one at
a time, but the encoder always reads them in document order.

\begin{figure}
    \centering
    \includegraphics[width=0.325\linewidth]{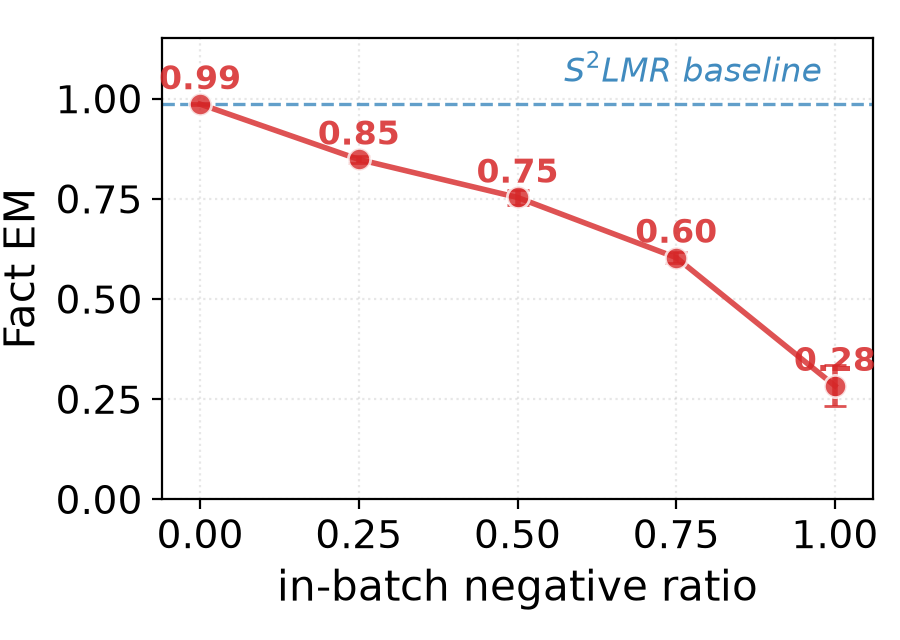}
    \hfill
    \includegraphics[width=0.325\linewidth]{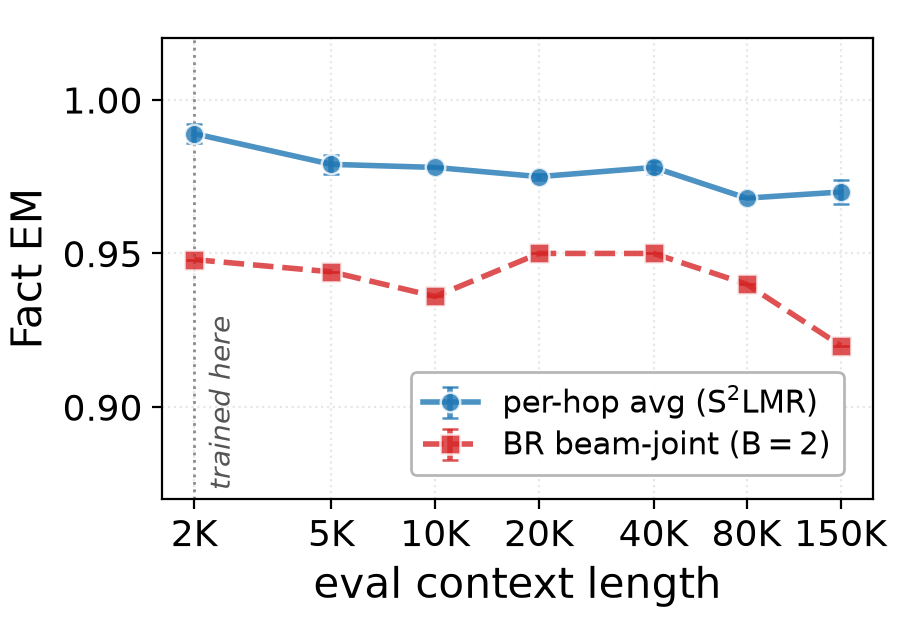}
    \hfill
    \includegraphics[width=0.325\linewidth]{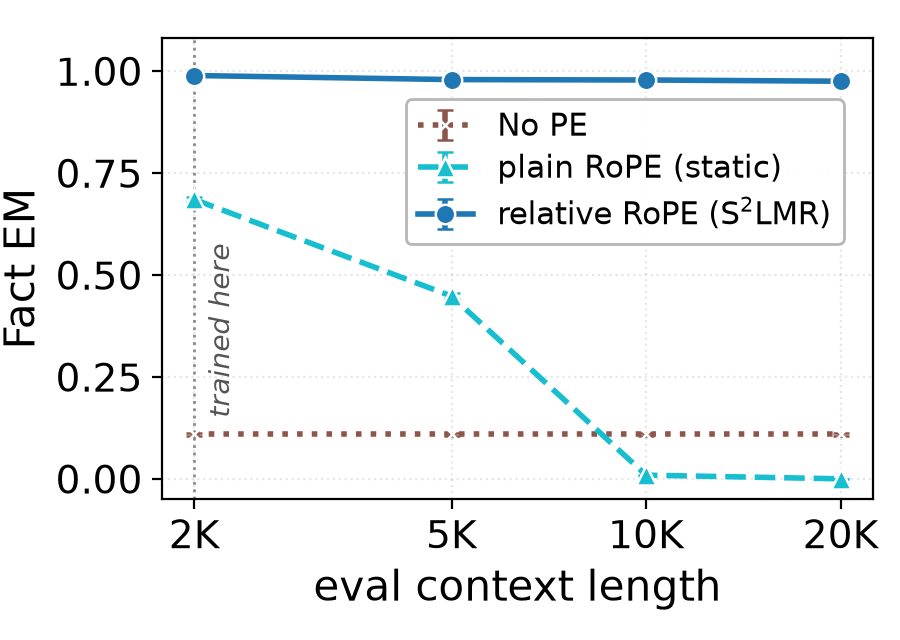}
    \caption{Verification of the three S$^2$LMR designs on BabiLong QA3, measured by Fact EM: \textbf{(a)} negative sampling with varying fractions of in-batch negatives; \textbf{(b)} teacher-forced versus beam-expanded per-hop supervision ($B{=}2$); and \textbf{(c)} re-anchored, static absolute, and no chunk-axis positional encoding. 
    Vertical dashed lines indicate the training context length (2K).}

    \label{fig:method_effect}
\end{figure}

During training, S$^2$LMR uses the gold retrieval prefix
$c^{(1)},\dots,c^{(t-1)}=(c_{i_1^\star},\dots,c_{i_{t-1}^\star})$ to construct
$s_t$. Rather than conditioning later-hop supervision on noisy model-generated
retrieval prefixes, S$^2$LMR therefore reformulates multi-hop retrieval as a
sequence of conditionally supervised retrieval decisions, where each hop
directly supervises the next evidence-selection decision given the preceding
gold evidence.
Fig.~\ref{fig:method_effect}b shows the effect. 
The performance gap between beam-expanded supervision and ours widens as context length increases.

The gold indices $(i_1^\star,\dots,i_K^\star)$ follow the annotated reasoning chain when available (MuSiQue, BabiLong) and document order otherwise (HotpotQA). 
Under this ordering, each hop has a single designated target, while later gold chunks remain candidates rather than being simultaneously treated as positive. 
This explicitly trains the model to distinguish which evidence should be retrieved next, rather than only which evidence belongs to the final support set.

\textbf{History-adaptive candidate positions.}
A candidate's position should describe where it lies relative to the evidence
retrieved so far, not where it lies in the document. Let
$i_{(1)}<\cdots<i_{(t-1)}$ be the sorted document indices of the chunks
retrieved before hop $t$, with $i_{(0)}=0$ and $i_{(t)}=N+1$. These indices
divide the remaining candidates into $t$ segments
$S_k=\{\,j \mid i_{(k)}<j<i_{(k+1)}\,\}$ for $k=0,\ldots,t-1$, one for each
gap between consecutive retrieved chunks. For $j\in S_k$ we set
\begin{align}
p_j^{(t)}
= k\Delta + 0.9\,\Delta\,\frac{j-i_{(k)}}{i_{(k+1)}-i_{(k)}},
\label{eq:pe}
\end{align}
with a fixed step size $\Delta>0$. Segment $k$ occupies
$[k\Delta,(k+0.9)\Delta)$, and the remaining $0.1\Delta$ separates it from
segment $k+1$. Two candidates on opposite sides of a retrieved chunk are
therefore assigned clearly different positions, even if they are adjacent in
the document.

The relative chunk axis is taken similarly to that of Q-RAG~\citep{sorokin2026q}. The main 
difference is that \rev{positions in S$^2$LMR are computed from the gold prefix in training.} This has two consequences. First, a position now tells the model
which two pieces of retrieved evidence a candidate lies between and how far
along that gap it sits, which is the information the next hop needs. Second,
$p_j^{(t)}<t\Delta\le K\Delta$ regardless of document length, so the model
never encounters a position at test time that it did not see during training.
Absolute chunk indices do not have this property because they grow with $N$.

Fig.~\ref{fig:method_effect}c compares three positional schemes: no positional encoding, static (absolute) RoPE, and our history-adaptive relative positions. Removing positional encoding entirely performs poorly across all context lengths. Static RoPE is competitive near the training length but degrades sharply beyond it, whereas ours remains accurate even at the longest context length we evaluated.

\textbf{Full-context contrastive supervision.}
Given $s_t$ and $p_j^{(t)}$, the score of candidate $c_j$ is given by 
\begin{align}
\operatorname{score}(s_t,c_j)
= \big\langle\,
f_\theta(s_t),\;
\operatorname{RoPE}\big(f_\theta(c_j),\,p_j^{(t)}\big)
\,\big\rangle,
\label{eq:score}
\end{align}
where RoPE~\citep{su2024roformer} is applied only to the candidate chunk embedding, while the query-state embedding $f_\theta(s_t)$ is kept at position zero. Since $p_j^{(t)}$ already measures the candidate's position relative to the retrieved history represented by $s_t$, this makes the score depend directly on the history-adaptive candidate position.

At training hop $t$, the candidate set contains every chunk not yet retrieved,
$\mathcal{A}_t=\mathcal{C}\setminus\{i_1^\star,\ldots,i_{t-1}^\star\}$, and
the loss is a softmax over this full set, averaged over hops:
\begin{align}
\mathcal{L}_t
= -\log
\frac{\exp\!\big(\operatorname{score}(s_t,c_{i_t^\star})/\tau\big)}
     {\sum_{j\in\mathcal{A}_t}\exp\!\big(\operatorname{score}(s_t,c_j)/\tau\big)},
\qquad
\mathcal{L}_{\mathrm{S^2LMR}}
= \frac{1}{K}\sum_{t=1}^{K}\mathcal{L}_t,
\label{eq:hop_nce}
\end{align}
with temperature $\tau$. We use neither sampled nor in-batch negatives. The
difficult negatives in multi-hop retrieval are chunks from the same context
that mention the right entities but are not the next step in the chain, and
they only enter the loss if the entire context is in the denominator.
Fig.~\ref{fig:method_effect}a shows the effect. Replacing same-context
negatives with in-batch negatives lowers Fact EM from $0.99$ to $0.28$, and
Table~\ref{tab:m1_evidence} shows that most of S$^2$LMR's margin over distractors
comes from same-context negatives. Because the gold sequence fixes the
training state at every hop, all $K$ decisions receive a direct gradient, and
no trajectory sampling or search is needed during training.

\textbf{Inference.}
At test time, the gold history is replaced by the model's own previous
retrievals, and the state construction, position rule, and scoring function
are unchanged. Chunk embeddings $\{f_\theta(c_j)\}_{j=1}^N$ are computed once
and cached. At hop $t$, the candidate set is
$\mathcal{A}_t=\mathcal{C}\setminus\{i_1,\ldots,i_{t-1}\}$. S$^2$LMR selects
$$i_t=\arg\max_{j\in\mathcal{A}_t}\operatorname{score}(s_t,c_j),$$ appends
$c_{i_t}$ to the history, and recomputes the candidate positions.  Algorithm~\ref{alg:ssr} in the appendix gives the full
procedure.

\subsection{Extensions of \texorpdfstring{S$^2$LMR}{S2LMR}}
\label{sec:relax}
The per-hop formulation extends naturally to settings with less information. We describe three
extensions: stopping without a known number of hops, training from unordered gold sets, and training
from answers only. Each reuses the state, positions and scoring function of Sec.~\ref{sec:ssr}, and
changes only how the number of hops or the training targets are obtained.

\textbf{Stopping without a known $K$.}
The number of hops a question needs is often unknown at test time. Since S$^2$LMR retrieves one chunk
per hop, stopping reduces to a decision before each hop $t$: take the next pick $c^{(t)}$, or return the
$t-1$ chunks retrieved so far. Let $\sigma_t=\operatorname{score}(s_t,c^{(t)})$ be the score of the next
pick and $\sigma'_t$ the second-highest score over $\mathcal{A}_t$. Besides a fixed budget of $T$ hops,
we consider two rules based on these scores: a score threshold, which stops once $\sigma_t<\kappa$, and
a learned stop, a logistic regression on
$\phi_t=(\sigma_t,\ \sigma_t-\sigma'_t,\ \sigma_t-\sigma_{t-1},\ \sigma_t-\sigma_1)$ and the hop index,
which predicts whether $c^{(1)},\dots,c^{(t-1)}$ already contain all gold chunks and stops when this
probability exceeds $\rho$. Both rules are fit post hoc on held-out data, so the retriever needs no
retraining.

\textbf{Training from unordered gold sets.}
Eq.~\ref{eq:hop_nce} uses an ordered gold sequence: hop $t$ targets $i_t^\star$, and its teacher-forced
state contains the gold chunks of earlier hops. Some datasets annotate only the set of supporting facts,
which is why Sec.~\ref{sec:ssr} falls back to document order on HotpotQA. Given only the gold set
$G=\{i_1^\star,\dots,i_K^\star\}$, let $\mathcal{L}_\pi$ be the loss of Eq.~\ref{eq:hop_nce} under
teacher forcing in order $\pi$, an ordering of $G$. We minimize the soft minimum over orderings,
\[
\mathcal{L}_{\text{soft}} = -\beta \log \sum_{\pi} \exp(-\mathcal{L}_\pi/\beta),
\]
with temperature $\beta>0$. As $\beta\to0$, $\mathcal{L}_{\text{soft}}$ approaches
$\min_\pi\mathcal{L}_\pi$, so each example is trained along the ordering the model currently finds
easiest rather than a fixed one. We enumerate all $K!$ orderings for $K\le3$ and sample $12$ for
$K\ge4$.

\textbf{Training from answers only.}
When only question--answer pairs are available, we let an instruction-tuned LLM label the gold chunks.
Given the question, the gold answer and the numbered context chunks, it lists the chunks needed to
derive the answer in reasoning order, starting from the chunk that links the question to an
intermediate entity. The listed chunks and their order replace $(i_1^\star,\dots,i_K^\star)$ in
Eq.~\ref{eq:hop_nce}, and training proceeds as before.

\section{Experiments}
\subsection{Experimental Setup}
We evaluate S$^2$LMR on both long-context retrieval and open-domain multi-hop retrieval, covering the primary settings of multi-step dense retrieval. 
Long-context evaluation is conducted on \textbf{RULER}~\citep{hsieh2024ruler}, which contains synthetic retrieval tasks with context lengths ranging from 4K to 1M tokens, including both single- and multi-step retrieval, and on \textbf{BabiLong}~\citep{kuratov2024babilong}, which evaluates multi-hop reasoning over contexts of up to millions of tokens. 
Open-domain evaluation is conducted on \textbf{HotpotQA}~\citep{yang2018hotpotqa} and \textbf{MuSiQue}~\citep{trivedi2022musique}, with typical context lengths of 1--2K and 2--3K tokens, respectively.

Following prior work, retrieval quality on HotpotQA and MuSiQue is evaluated using Supporting Fact F1/EM together with final answer F1/EM. For RULER and BabiLong, we report Fact EM across retrieval subtasks and context lengths. Unless otherwise specified, we use identical retrieval budgets and decoding procedures where applicable. \rev{Q-RAG results come from the authors' released checkpoints or from retraining with the official code.
Appendix~\ref{app:qrag_protocol} details the sources and compares both methods under alternative evaluation protocols, including those used in the Q-RAG paper. Chunk-axis PE is disabled on RULER, HotpotQA and MuSiQue, where evidence comes from independent passages and its position carries no information about the next hop.} Detailed training hyperparameters are listed in Appendix~\ref{app:hparams}.
In the result tables, $\circ$ marks \emph{reported} baselines whose scores are taken directly from the original papers, while $\checkmark$ marks \emph{reproduced} baselines that we fine-tuned and/or evaluated using released code or publicly available checkpoints.

\begin{figure}
    \centering
    \includegraphics[width=1\linewidth]{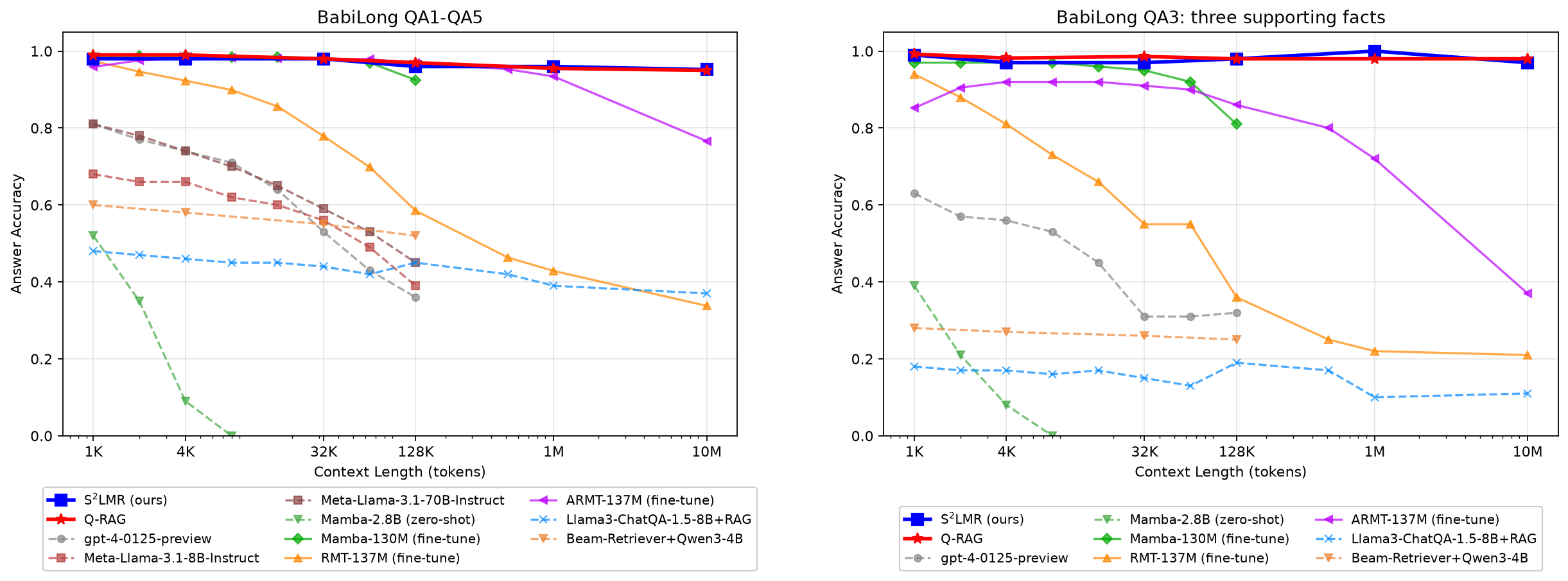}
    \caption{Comparison of answer accuracy on the long-context benchmark BabiLong. Solid lines denote methods fine-tuned on the BabiLong, while dashed lines denote zero-shot methods. a) Average
    performance across tasks Q1–QA5. b) Performance on the hardest task, QA3, which requires the
    longest reasoning chain and temporal awareness.}
    \label{fig:babilong}
\end{figure}

\subsection{Retrieval Performance}

\textbf{1) Ultra-long Context Reasoning.} 
We compare S$^2$LMR with  the state-of-the-art long-context retrieval methods including Q-RAG~\citep{sorokin2026q}, Beam Retriever~\citep{zhang2024end}, ARMT~\citep{rodkin2024associative}, RMT~\citep{bulatov2022recurrent}, Mamba2~\citep{dao2024transformers}, and other long-context language models and LLM-based agents~\citep{achiam2023gpt,grattafiori2024llama} reported in prior work.  
Figure~\ref{fig:babilong} reports average performance across all QA tasks. S$^2$LMR maintains strong performance across the entire context range, remaining above 95\% even at 10M tokens and closely matching Q-RAG throughout. It also substantially outperforms existing long-context architectures, including recurrent transformers and state-space models.

We further examine QA3, the most challenging BabiLong task requiring three-step reasoning. As shown in Figure~\ref{fig:babilong}, S$^2$LMR again maintains strong performance as context length increases, closely matching Q-RAG across the full range. Together, these results show that a supervised bi-encoder can effectively support multi-step retrieval and reasoning even in extremely long contexts.

\begin{table}[tb]
\centering
\small
\caption{Results on the RULER benchmark, evaluating long-context retrieval performance across various context lengths. \textbf{S} (Single-needle): Find one value for one key. \textbf{MK} (Multi-keys): Find one value for one key among many. \textbf{MV} (Multi-values): Find all values for one key. \textbf{MQ} (Multi-query): Answer multiple questions over the context. \textbf{MH QA}: open domain multi-hop question answering. \textbf{SH QA}: single-hop question answering.}
\label{tab:ruler}
\resizebox{\columnwidth}{!}{%
\begin{tabular}{ll ccc ccc c c c cc}
\toprule
& & \multicolumn{3}{c}{S} & \multicolumn{3}{c}{MK} & MV & MQ & NIAH & \multicolumn{2}{c}{QA} \\
\cmidrule(lr){3-5} \cmidrule(lr){6-8} \cmidrule(lr){12-13}
Len & Methods & 1-st & 2-nd & 3-rd & 1-st & 2-nd & 3-rd & & & Avg. & SH & MH \\
\midrule
\multirow{7}{*}{4K}
 & $^{\circ}$Titans          & 98.4 & 99.8 & 89.4 & n/a & n/a & n/a & n/a  & n/a  & n/a  & n/a  & n/a \\
 & $^{\circ}$Atlas           & 99.2 & 100  & 90.6 & n/a & n/a & n/a & n/a  & n/a  & n/a  & n/a  & n/a \\
 & $^{\circ}$Mamba2-Hybrid   & 100  & 100  & 95.7 & 89.5 & 95.5 & 96 & 97.9 & 97.6 & 96.5 & 56.5 & 48.8 \\
 & $^{\circ}$LongRoPE2-8B    & 100  & 100  & 99   & 100  & 100  & 97 & 99.7 & 100  & 99.7 & \textbf{79}   & 60 \\
 & $^{\checkmark}$Beam-Retriever & 100 & 98 & 98 & 98  & 98   & 97 & 98   & 99   & 98.5 & 29.0 & 39.0 \\
 & $^{\checkmark}$Q-RAG                     & 100  & 100  & 100  & 100  & 100  & 100 & 100  & 100  & 100  & 55   & \textbf{64} \\
 & \textbf{S$^2$LMR (Ours)}              & 100  & 100  & 100  & 100  & 100  & 100 & 100  & 100  & 100  & 56   & \textbf{64} \\
\midrule
\multirow{7}{*}{16K}
 & $^{\circ}$Titans          & 96.2 & 80.2 & n/a & n/a & n/a & n/a & n/a  & n/a  & n/a  & n/a  & n/a \\
 & $^{\circ}$Atlas           & 97   & 84   & n/a & n/a & n/a & n/a & n/a  & n/a  & n/a  & n/a  & n/a \\
 & $^{\circ}$Mamba2-Hybrid   & 100  & 100  & 81.5 & 92  & 92.2 & 83 & 89.8 & 90.2 & 91.1 & 48.8 & 44 \\
 & $^{\circ}$LongRoPE2-8B    & 100  & 100  & 99   & 99  & 100  & 98 & 95   & 98.2 & 98.8 & \textbf{69}   & 58 \\
 & $^{\checkmark}$Beam-Retriever & 100 & 100 & 97 & 96.5 & 96 & 95 & 80   & 98   & 95.3 & 24.0 & 35.0 \\
 & $^{\checkmark}$Q-RAG                     & 100  & 100  & 100  & 100  & 100  & 100 & 100  & 100  & 100  & 53   & 60 \\
 & \textbf{S$^2$LMR (Ours)}             & 100  & 100  & 100  & 100  & 100  & 100 & 100  & 100  & 100  & 51   & \textbf{62} \\
\midrule
\multirow{4}{*}{32K}
 & $^{\circ}$Mamba2-Hybrid   & 100  & 100  & 96.7 & 84  & 76.5 & 81.5 & 84.3 & 80.9 & 88.0 & 41.8 & 38.5 \\
 & $^{\circ}$LongRoPE2-8B    & 100  & 100  & 100   & 99  & 98   & 100   & 98 & 96.2   & 98.9 & \textbf{72}   & 55 \\
 & $^{\checkmark}$Q-RAG                     & 100  & 100  & 100  & 100  & 100  & 100 & 100  & 100  & 100  & 52   & \textbf{63} \\
 & \textbf{S$^2$LMR (Ours)}             & 100  & 100  & 100  & 100  & 100  & 100 & 100  & 100  & 100  & 50   & \textbf{63} \\
\midrule
\multirow{4}{*}{128K}
 & $^{\checkmark}$Mamba2-Hybrid   & 100  & 92   & 96   & 81   & 81   & 75   & 48.8 & 72.5 & 80.8 & 36   & 26 \\

 & $^{\circ}$LongRoPE2-8B    & 100  & 100  & 99   & 96  & 91   & 94   & 96.5 & 97   & 96.7 & \textbf{56}   & 50 \\
 & $^{\checkmark}$Q-RAG                     & 100  & 100  & 100  & 100  & 100  & 100 & 100  & 100  & 100  & 50   & 61 \\
 & \textbf{S$^2$LMR (Ours)}             & 100  & 100  & 100  & 100  & 100  & 100 & 100  & 100  & 100  & 47   & \textbf{62} \\
\midrule
\multirow{2}{*}{1M}
 & $^{\checkmark}$Q-RAG                     & 100  & 100  & 100 & 98.5 & 99.0 & 100 & 100  & 100  & 99.7  & \textbf{50}   & 56 \\
 & \textbf{S$^2$LMR (Ours)}             & 100  & 100  & 100  & \textbf{100} & 99.0 & 100 & 100  & 100  & \textbf{99.8} & 46   & \textbf{59} \\
\bottomrule
\end{tabular}%
}
\end{table}


\textbf{2) Long-context retrieval.} 
Table~\ref{tab:ruler} reports results on RULER across context lengths from 4K to 1M tokens. 
The  baselines include Beam Retriever~\citep{zhang2024end}, Titans~\citep{behrouz2026titans}, Atlas~\citep{behrouz2025atlas}, Mamba2~\citep{dao2024transformers}, LongRoPE2~\citep{shang2025longrope2}, and Q-RAG~\citep{sorokin2026q}.

S$^2$LMR consistently matches or exceeds Q-RAG on nearly every retrieval task across the full context range. 
On single-needle, multi-key, multi-value, and multi-query retrieval, both methods achieve near-perfect accuracy, while S$^2$LMR also remains competitive with Q-RAG on long-context question answering and substantially outperforms previous long-context architectures, including recurrent transformers and state-space models. 
These results demonstrate that S$^2$LMR remains effective across diverse retrieval tasks and context lengths, including contexts of up to 1M tokens.
\begin{table}[t]
\centering
\caption{Comparison on HotPotQA (1--2K tokens) and MuSiQue (2--3K tokens). Bold/underline mark best/2nd.}
\label{tab:hotpot_musique}
\resizebox{\textwidth}{!}{%
\begin{tabular}{@{}l cccc cccc cc@{}}
\toprule
& \multicolumn{4}{c}{\textbf{HotPotQA}} & \multicolumn{4}{c}{\textbf{MuSiQue}} & \multicolumn{2}{c}{\textbf{Avg}} \\
\cmidrule(lr){2-5} \cmidrule(lr){6-9} \cmidrule(lr){10-11}
\textbf{Methods} & Fact F1 & Fact EM & Ans F1 & Ans EM & Fact F1 & Fact EM & Ans F1 & Ans EM & Ans F1 & Ans EM \\
\midrule
\multicolumn{11}{c}{Finetuned} \\
\midrule
$^{\checkmark}$Q-RAG            & 0.92             & 0.87             & 0.76             & 0.59             & 0.78             & 0.57             & 0.55             & 0.41             & 0.66             & \underline{0.50} \\
$^{\circ}$Beam-Retriever        & \textbf{0.97}    & \textbf{0.94}    & \underline{0.77} & \underline{0.61} & \textbf{0.86}    & \textbf{0.69}    & \textbf{0.59}    & \textbf{0.43}    & \textbf{0.68}    & \textbf{0.52} \\
$^{\circ}$Search-R1             & 0.81             & 0.66             & 0.65             & 0.52             & --               & --               & --               & --               & --               & -- \\
$^{\circ}$RAG-RL                & 0.82             & --               & 0.69             & --               & --               & --               & --               & --               & --               & -- \\
\textbf{S$^2$LMR (Ours)}        & \underline{0.95} & \underline{0.90} & \textbf{0.78}    & \textbf{0.62}    & \underline{0.83} & \underline{0.67} & \underline{0.56} & \underline{0.42} & \underline{0.67} & \textbf{0.52} \\
\midrule
\multicolumn{11}{c}{Zero Shot} \\
\midrule
$^{\circ}$Multi-step RAG        & 0.73             & 0.54             & 0.65             & 0.50             & 0.51             & 0.30             & 0.40             & 0.27             & 0.53             & 0.39 \\
$^{\circ}$GraphReader           & --               & --               & 0.46             & 0.24             & --               & --               & 0.40             & 0.20             & 0.43             & 0.22 \\
$^{\circ}$Single step RAG       & --               & --               & 0.53             & 0.39             & --               & --               & 0.28             & 0.17             & 0.41             & 0.28 \\
\bottomrule
\end{tabular}%
}
\end{table}

\textbf{3) Open-domain multi-hop retrieval.}
Table~\ref{tab:hotpot_musique} compares S$^2$LMR with multiple strong baselines on HotPotQA and MuSiQue, including Beam Retriever~\citep{zhang2024end}, Q-RAG~\citep{sorokin2026q}, Search-R1~\citep{jin2025search}, RAG-RL~\citep{huang2025rag}, GraphReader~\citep{li2024graphreader}, and a conventional single-step RAG baseline~\citep{lewis2020retrieval}. On HotPotQA, S$^2$LMR achieves the best downstream answer performance (Ans F1/EM) while remaining competitive with Beam Retriever on supporting-fact retrieval. Across both benchmarks, S$^2$LMR outperforms Q-RAG on every metric and ties Beam Retriever for the best average answer EM. Notably, S$^2$LMR fine-tunes only the dense retriever rather than the language model, yet achieves comparable downstream performance. These results show that S$^2$LMR generalizes beyond long-context retrieval to standard short-context multi-hop QA. Q-RAG's own MuSiQue evaluation stops once all gold facts are retrieved. Appendix~\ref{app:qrag_protocol} compares both methods under this protocol as well.

\subsection{Training Stability}
Training efficiency and stability directly affect the practical cost and reproducibility of retriever training. We compare S$^2$LMR with Q-RAG on BabiLong QA3 from two complementary perspectives: (1) robustness to hyperparameter choices and (2) stability of training dynamics. 

\begin{figure}
    \centering
    \begin{subfigure}[t]{0.49\linewidth}
        \centering
        \includegraphics[width=\linewidth]{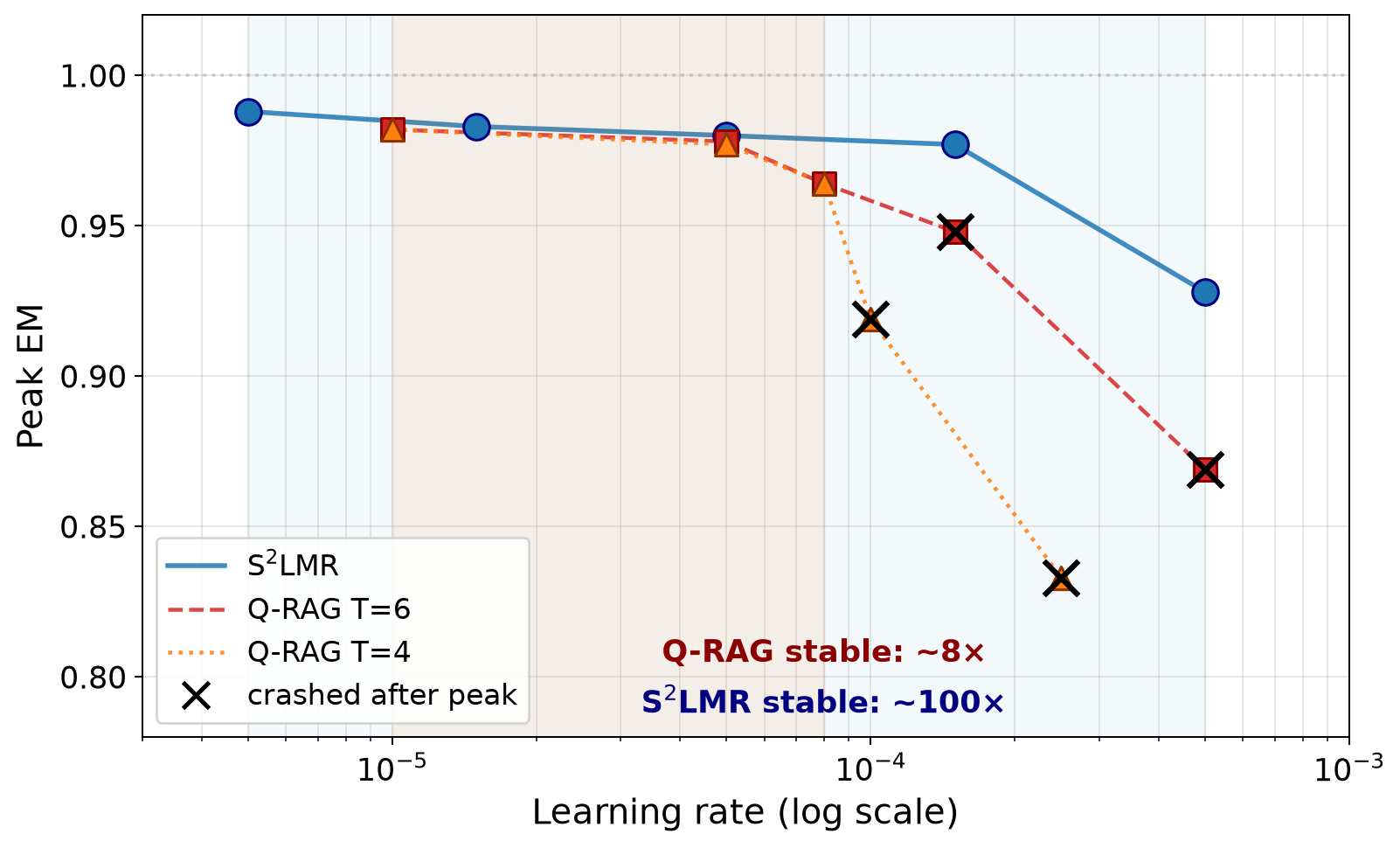}
        \caption{Learning-rate robustness.}
        \label{fig:lr_range}
    \end{subfigure}
    \hfill
    \begin{subfigure}[t]{0.49\linewidth}
        \centering
        \includegraphics[width=\linewidth]{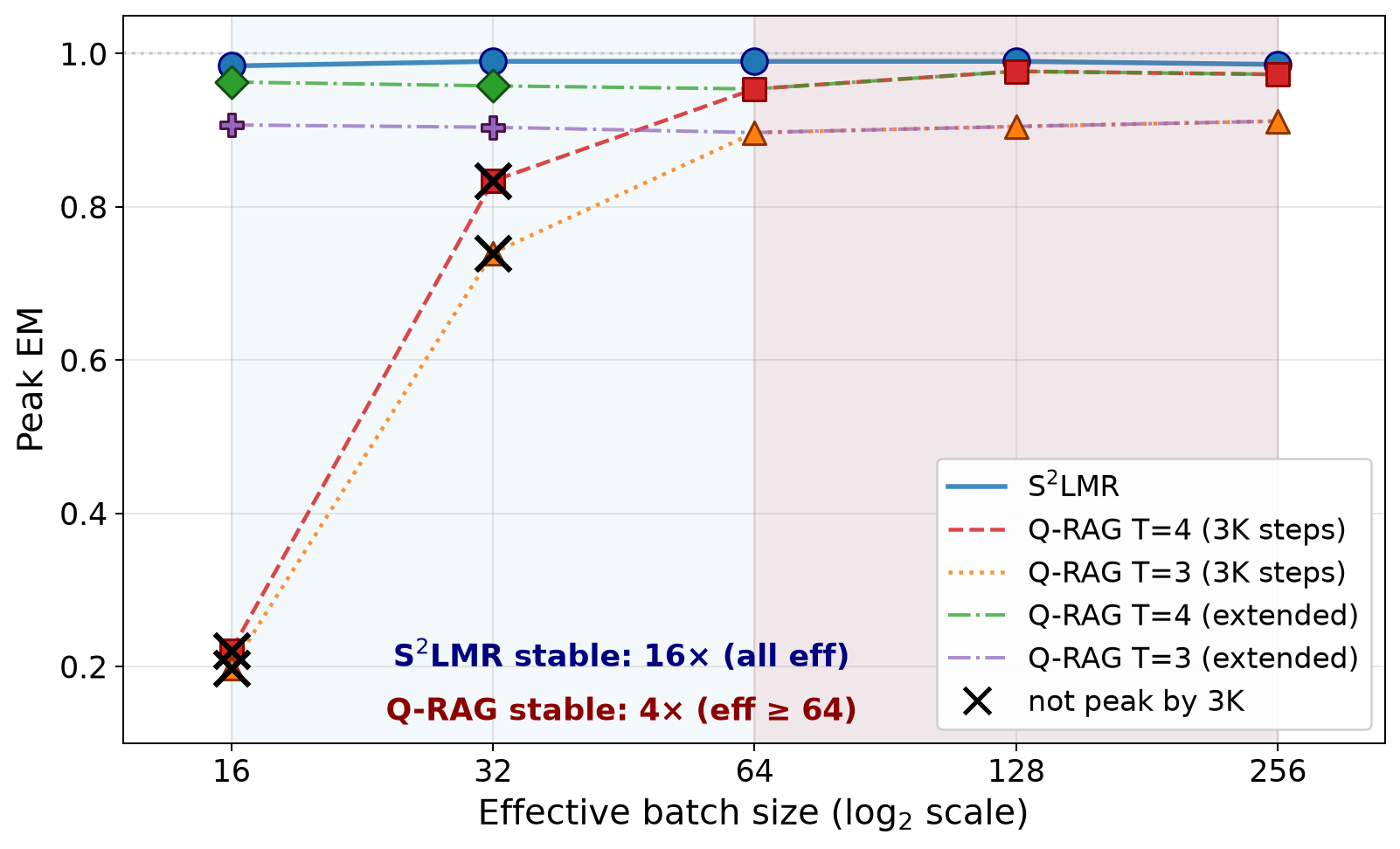}
        \caption{Batch-size robustness.}
        \label{fig:batch_range}
    \end{subfigure}
    \caption{Hyperparameter robustness on BabiLong QA3. Shaded regions indicate the stable training range; black $\times$ marks Q-function collapse.}
    \label{fig:hparam_range}
\end{figure}

\label{sec:exp:stability:hp} \textbf{1) Hyperparameter robustness.} Figure~\ref{fig:hparam_range} sweeps S$^2$LMR and Q-RAG over learning rate (panel a) and effective batch size (panel b), while keeping the trainable retriever backbone fixed.
\textbf{\em Learning-rate robustness.} S$^2$LMR maintains near-peak performance across a learning-rate range spanning nearly two orders of magnitude ($\approx 100\times$). Q-RAG remains stable only within a much narrower $\approx 8\times$ range, with its Q-function collapsing at higher learning rates (crashes marked by $\times$). Thus, S$^2$LMR provides a usable learning-rate range more than an order of magnitude wider than Q-RAG's.
\textbf{\em Batch-size robustness.} At a fixed 3K-step training budget, S$^2$LMR reaches near-peak EM across a $16\times$ batch-size range, from 16 to 256. Under the same budget, Q-RAG has a usable range of only $4\times$: at batch sizes below 64, it fails to reach peak performance within the standard budget and requires several times more optimizer steps to escape its initial low-EM phase. This additional step requirement further increases its practical training cost.

Beyond robustness to individual hyperparameters, S$^2$LMR also has fewer method-specific hyperparameters to tune. Q-RAG's value-based formulation introduces additional TD-related choices, including the discount factor, target-network update schedule, and value-network learning rate, which are absent from the supervised bi-encoder formulation.

\begin{figure}[t]
    \centering
    \begin{subfigure}[t]{0.325\linewidth}
        \centering
        \includegraphics[width=\linewidth]{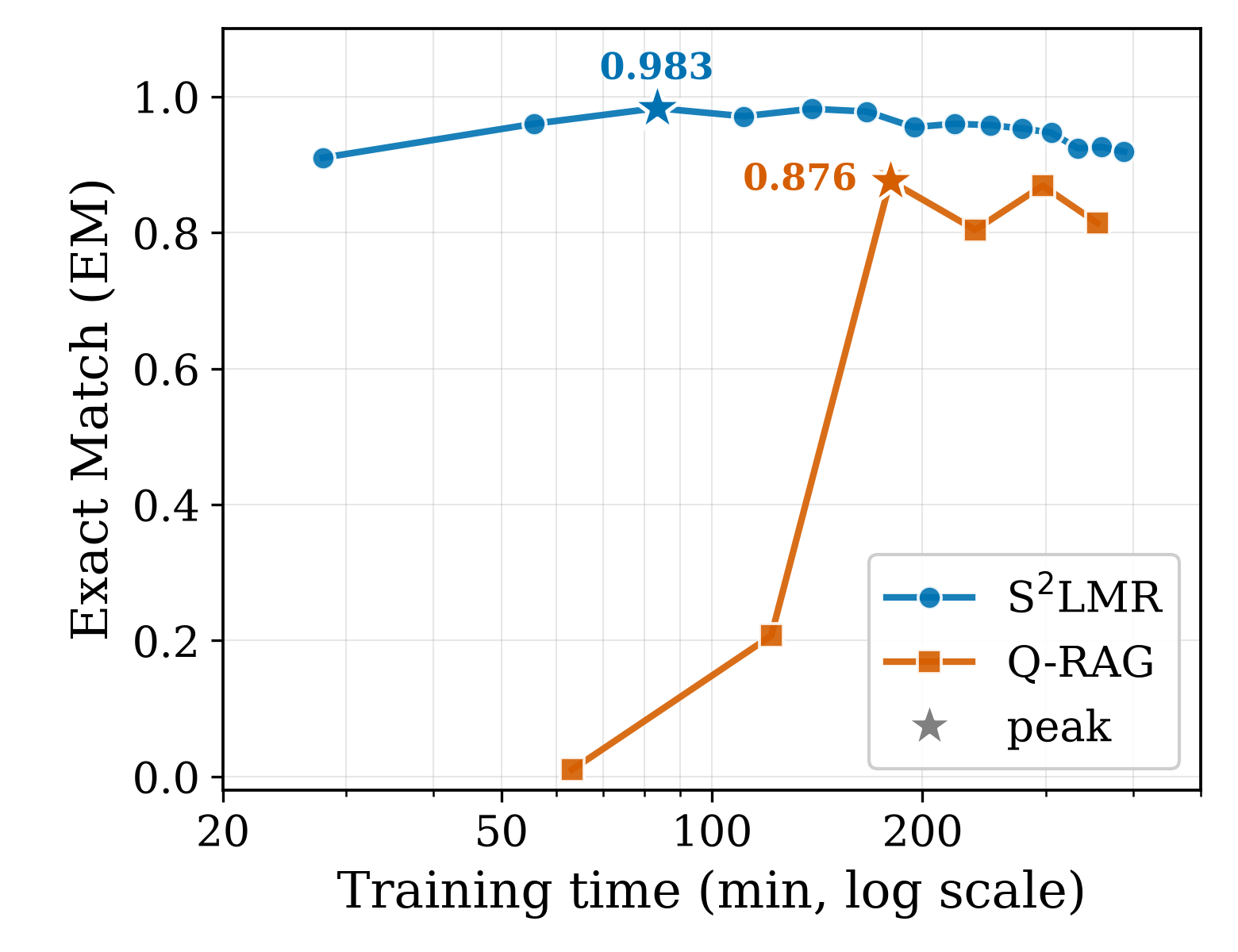}
        \caption{Peak-arrival speed.}
        \label{fig:stability:peak}
    \end{subfigure}
    \hfill
    \begin{subfigure}[t]{0.325\linewidth}
        \centering
        \includegraphics[width=\linewidth]{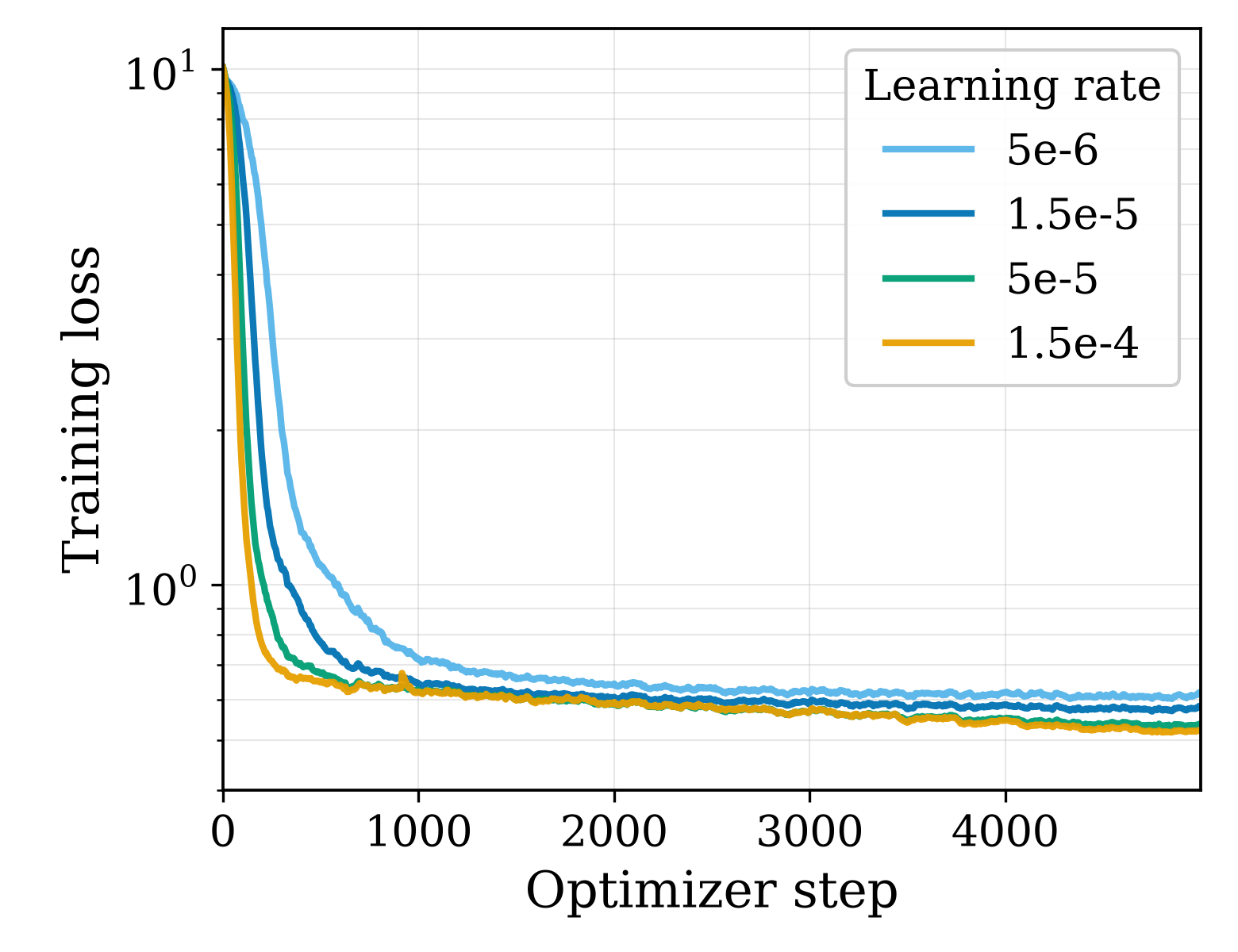}
        \caption{S$^2$LMR training loss.}
        \label{fig:stability:ssr}
    \end{subfigure}
    \hfill
    \begin{subfigure}[t]{0.325\linewidth}
        \centering
        \includegraphics[width=\linewidth]{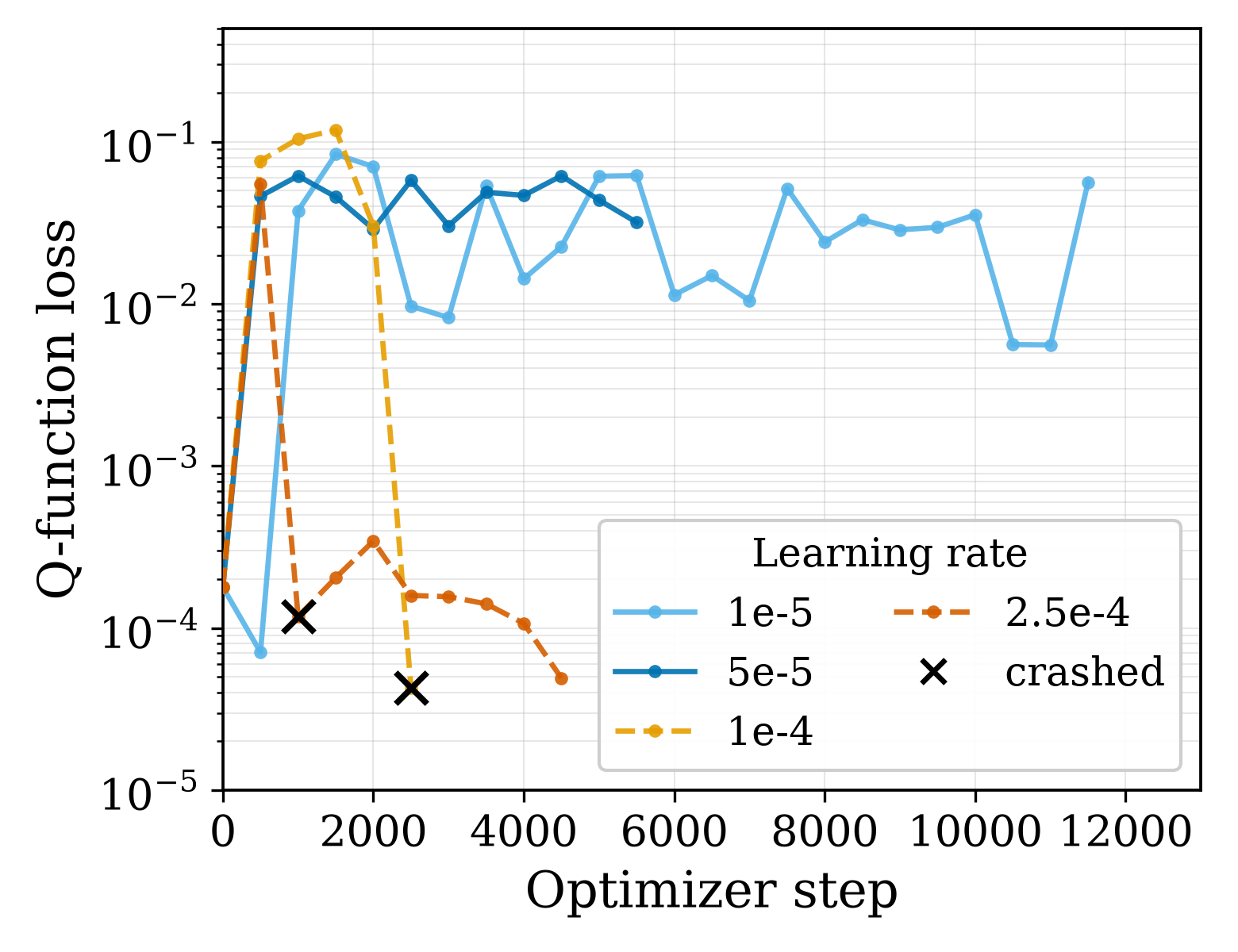}
        \caption{Q-RAG qf-loss.}
        \label{fig:stability:qrag}
    \end{subfigure}
    \caption{Training stability on BabiLong QA3.
(\subref{fig:stability:peak})~Peak EM vs.\ training time.
(\subref{fig:stability:ssr})~S$^2$LMR training loss across learning rates.
(\subref{fig:stability:qrag})~Q-RAG Q-function loss across learning rates.
}
    \label{fig:stability}
\end{figure}
\textbf{2) Training dynamics.} \label{sec:exp:stability:dynamics} Figure~\ref{fig:stability} compares the training dynamics of S$^2$LMR and Q-RAG at the same learning rate, using Q-RAG's reported best setting.
\textbf{\em Peak-arrival speed.}
S$^2$LMR reaches its peak EM of $0.983$ within $\sim\!83$ minutes of wall-clock training, whereas Q-RAG requires $\sim\!180$ minutes to reach a peak of $0.876$ (Figure~\ref{fig:stability}a). 
S$^2$LMR converges $2.2\times$ faster and to $+0.11$ higher final EM.
\textbf{\em Loss stability.} Across the learning rates within each method's stable range, S$^2$LMR's training loss decreases smoothly, whereas Q-RAG's Q-function loss is substantially more volatile and diverges at larger learning rates (Figure~\ref{fig:stability}b,c). The smoother optimization of S$^2$LMR also makes its performance less sensitive to the learning rate within the stable range, reducing the need for repeated restarts or fine-grained tuning.

\subsection{Training Efficiency}

\begin{wraptable}[15]{r}{0.62\columnwidth}
\vspace{-0.8cm}
\centering
\small
\setlength{\tabcolsep}{3pt}
\caption{Time cost on BabiLong QA3.}
\label{tab:time_cost}
\begin{tabular}{@{}lcccccc@{}}
\toprule
\multirow{2}{*}{\textbf{Method}} & \multirow{2}{*}{\textbf{Train (h)}} & \multicolumn{5}{c}{\textbf{Infer.\ (s/sample)}} \\
\cmidrule(l){3-7}
 & & 1K & 32K & 128K & 1M & 10M \\
\midrule
\multicolumn{7}{l}{\emph{Retrieval-only (+0.44s Qwen3-4B)}} \\
S$^2$LMR            & \textbf{2.25} & 0.20 & 1.00 & 3.20 & 27.8 & 279 \\
Q-RAG               & 54.4          & 0.20 & 1.10 & 3.42 & 29.0 & 318 \\
Beam Retriever      & $\sim$450     & 1.2  & 39   & 155  & $\sim$1200 & $\sim$12000 \\
\midrule
\multicolumn{7}{l}{\emph{End-to-end fine-tuned}} \\
RMT-137M            & $\sim$20      & 0.21 & 0.90 & 3.15 & 24.5 & 253 \\
Llama3-8B+RAG       & $>$40         & 0.65 & 0.75 & 1.23 & 7.82 & 64.8 \\
Mamba-130M          & $>$400        & 4.21 & 81.8 & OOM  & OOM  & OOM  \\
\midrule
\multicolumn{7}{l}{\emph{End-to-end zero-shot}} \\
Mamba-2.8B          & 0             & 8.81 & 133  & OOM  & OOM   & OOM   \\
Llama-3.1-8B        & 0             & 0.61 & 11.9 & 49   & limit & limit \\
\bottomrule
\end{tabular}
\end{wraptable}
We compare the training and inference costs of S$^2$LMR and representative baselines on a single NVIDIA RTX A6000 GPU (Table~\ref{tab:time_cost}).
For retrieval-based methods, we report retrieval time and add $0.44$\,s for Qwen3-4B answer generation; end-to-end LLM methods report total inference time.
``OOM'' denotes out-of-memory runs, and ``limit'' denotes runs exceeding the model's maximum context length.
For methods without reported training compute, we estimate training time from reported FLOPs or training data size and calibrate it on our hardware.
On BabiLong QA3, S$^2$LMR requires only $2.25$ hours of training, compared with $54.4$ hours for our Q-RAG reproduction, corresponding to a roughly $24\times$ lower training cost. A similar cost gap is observed on RULER. Compared with methods that fine-tune long-context language models, which can require tens to thousands of GPU-hours, S$^2$LMR achieves comparable or stronger retrieval performance at substantially lower training cost.

\subsection{Evaluation on Extended Settings}
\label{sec:supervision}
We evaluate the three extensions of \texorpdfstring{S$^2$LMR}{S2LMR} (see Sec.~\ref{sec:relax}) on multi-hop QA.

\begin{table}[t]
\centering
\small
\begin{minipage}[b]{0.50\linewidth}
\centering
\caption{Retrieval rules on MuSiQue (full dev) when $K$ is unknown.}
\label{tab:stopping}
\end{minipage}\hfill
\begin{minipage}[b]{0.46\linewidth}
\centering
\caption{Cross-dataset Fact EM: each model is trained on the other dataset.}
\label{tab:ood}
\end{minipage}
\par\vspace{2pt}
\begin{minipage}[t]{0.50\linewidth}
\centering
\begin{tabular}{@{}lcc@{}}
\toprule
Retrieval rule & Fact EM & \#picks \\[\dimexpr\aboverulesep+\belowrulesep+\cmidrulewidth\relax]
\midrule
Known $K=|\text{gold}|$ & 0.67 & 2.65 \\
Fixed $T{=}2/3/4$ & 0.39/0.65/0.77 & 2/3/4 \\
Score threshold & 0.63 & 2.60 \\
Learned stop & 0.65 & 2.70 \\
\bottomrule
\end{tabular}
\end{minipage}\hfill
\begin{minipage}[t]{0.46\linewidth}
\centering
\begin{tabular}{@{}lcc@{}}
\toprule
& \multicolumn{2}{c}{Evaluated on} \\
\cmidrule(l){2-3}
Method & MuSiQue & HotpotQA \\
\midrule
Q-RAG & 0.55 & 0.68 \\
S$^2$LMR & 0.54 & 0.76 \\
S$^2$LMR, order-agnostic & \textbf{0.60} & \textbf{0.77} \\
\bottomrule
\end{tabular}
\end{minipage}
\end{table}

\textbf{Stopping without a known $K$.}
We evaluate on MuSiQue, whose questions need 2--4 paragraphs, and calibrate the stopping rules by 2-fold
cross-validation on dev. Table~\ref{tab:stopping} shows a trade-off between recall and compactness.
A larger fixed budget raises recall: with $T=4$, Fact EM rises to 0.77, and answer quality also improves
(QwQ-32B Ans F1/EM 0.62/0.47 vs.\ 0.56/0.42 with known $K$), suggesting that the reader is robust to a
few extra paragraphs. The learned stop instead keeps the evidence set compact, recovering most of the
known-$K$ Fact EM (0.65 vs.\ 0.67) with nearly the same number of picks (2.70 vs.\ 2.65).

\textbf{Training from unordered gold sets.}
Trained without order labels, $\mathcal{L}_{\text{soft}}$ performs on par with the fixed-order model
in-domain (Fact EM 0.90 vs.\ 0.90 on HotpotQA, 0.69 vs.\ 0.67 on MuSiQue) and transfers better across
datasets. Table~\ref{tab:ood} trains on one dataset and evaluates on the other, following Q-RAG's
protocol (up to four steps on MuSiQue, stopping once all gold paragraphs are found; two steps on
HotpotQA). The fixed-order model already outperforms Q-RAG from MuSiQue to HotpotQA (0.76 vs.\ 0.68)
and is on par in the other direction (0.54 vs.\ 0.55), while the order-agnostic model performs best in
both directions.

\textbf{Training from answers only.}
On HotpotQA, we discard all supporting-fact annotations and generate pseudo-labels with Qwen3-8B; the
annotations are used only for evaluation. The pseudo-labels identify both gold paragraphs for 77\% of
questions. Trained on them, S$^2$LMR reaches 0.88 Fact EM on the full dev set, close to the 0.90
obtained with gold labels and well above the 0.65 of the zero-shot encoder.

\section{Conclusion}
\label{sec:conclusion}

We introduced S$^2$LMR, a simple supervised bi-encoder for long-context multi-step retrieval.
We showed that supervised retrievers fall behind in this setting not because supervision is too weak, but because their training is misaligned with how they are used.
By correcting these misalignments, S$^2$LMR matches or outperforms the strongest RL-based retriever at up to 10M tokens, while training substantially faster and remaining robust to hyperparameters.
Future work includes extending S$^2$LMR to heterogeneous retrieval corpora with evidence distributed across different sources. 

\bibliography{iclr2027_conference}

\begin{thebibliography}{42}
\providecommand{\natexlab}[1]{#1}
\providecommand{\url}[1]{\texttt{#1}}
\expandafter\ifx\csname urlstyle\endcsname\relax
  \providecommand{\doi}[1]{doi: #1}\else
  \providecommand{\doi}{doi: \begingroup \urlstyle{rm}\Url}\fi

\bibitem[Achiam et~al.(2023)Achiam, Adler, Agarwal, Ahmad, Akkaya, Aleman, Almeida, Altenschmidt, Altman, Anadkat, et~al.]{achiam2023gpt}
Josh Achiam, Steven Adler, Sandhini Agarwal, Lama Ahmad, Ilge Akkaya, Florencia~Leoni Aleman, Diogo Almeida, Janko Altenschmidt, Sam Altman, Shyamal Anadkat, et~al.
\newblock Gpt-4 technical report.
\newblock \emph{arXiv preprint arXiv:2303.08774}, 2023.

\bibitem[Anokhin et~al.(2024)Anokhin, Semenov, Sorokin, Evseev, Kravchenko, Burtsev, and Burnaev]{anokhin2024arigraph}
Petr Anokhin, Nikita Semenov, Artyom Sorokin, Dmitry Evseev, Andrey Kravchenko, Mikhail Burtsev, and Evgeny Burnaev.
\newblock Arigraph: Learning knowledge graph world models with episodic memory for llm agents.
\newblock \emph{arXiv preprint arXiv:2407.04363}, 2024.

\bibitem[Asai et~al.(2024)Asai, Wu, Wang, Sil, and Hajishirzi]{asai2024self}
Akari Asai, Zeqiu Wu, Yizhong Wang, Avi Sil, and Hannaneh Hajishirzi.
\newblock Self-rag: Learning to retrieve, generate, and critique through self-reflection.
\newblock In \emph{International conference on learning representations}, volume 2024, pp.\  9112--9141, 2024.

\bibitem[Behrouz et~al.(2025)Behrouz, Li, Kacham, Daliri, Deng, Zhong, Razaviyayn, and Mirrokni]{behrouz2025atlas}
Ali Behrouz, Zeman Li, Praneeth Kacham, Majid Daliri, Yuan Deng, Peilin Zhong, Meisam Razaviyayn, and Vahab Mirrokni.
\newblock Atlas: Learning to optimally memorize the context at test time.
\newblock \emph{arXiv preprint arXiv:2505.23735}, 2025.

\bibitem[Behrouz et~al.(2026)Behrouz, Zhong, and Mirrokni]{behrouz2026titans}
Ali Behrouz, Peilin Zhong, and Vahab Mirrokni.
\newblock Titans: Learning to memorize at test time.
\newblock \emph{Advances in Neural Information Processing Systems}, 38:\penalty0 113506--113543, 2026.

\bibitem[Bulatov et~al.(2022)Bulatov, Kuratov, and Burtsev]{bulatov2022recurrent}
Aydar Bulatov, Yury Kuratov, and Mikhail Burtsev.
\newblock Recurrent memory transformer.
\newblock \emph{Advances in neural information processing systems}, 35:\penalty0 11079--11091, 2022.

\bibitem[Dao \& Gu(2024)Dao and Gu]{dao2024transformers}
Tri Dao and Albert Gu.
\newblock Transformers are ssms: Generalized models and efficient algorithms through structured state space duality.
\newblock \emph{arXiv preprint arXiv:2405.21060}, 2024.

\bibitem[Das et~al.(2019)Das, Dhuliawala, Zaheer, and McCallum]{das2019multi}
Rajarshi Das, Shehzaad Dhuliawala, Manzil Zaheer, and Andrew McCallum.
\newblock Multi-step retriever-reader interaction for scalable open-domain question answering.
\newblock \emph{arXiv preprint arXiv:1905.05733}, 2019.

\bibitem[Gao et~al.(2023)Gao, Xiong, Gao, Jia, Pan, Bi, Dai, Sun, Wang, and Wang]{gao2023retrieval}
Yunfan Gao, Yun Xiong, Xinyu Gao, Kangxiang Jia, Jinliu Pan, Yuxi Bi, Yi~Dai, Jiawei Sun, Meng Wang, and Haofen Wang.
\newblock Retrieval-augmented generation for large language models: A survey.
\newblock \emph{arXiv preprint arXiv:2312.10997}, 2023.

\bibitem[Grattafiori et~al.(2024)Grattafiori, Dubey, Jauhri, Pandey, Kadian, Al-Dahle, Letman, Mathur, Schelten, Vaughan, et~al.]{grattafiori2024llama}
Aaron Grattafiori, Abhimanyu Dubey, Abhinav Jauhri, Abhinav Pandey, Abhishek Kadian, Ahmad Al-Dahle, Aiesha Letman, Akhil Mathur, Alan Schelten, Alex Vaughan, et~al.
\newblock The llama 3 herd of models.
\newblock \emph{arXiv preprint arXiv:2407.21783}, 2024.

\bibitem[Gu \& Dao(2023)Gu and Dao]{gu2023mamba}
Albert Gu and Tri Dao.
\newblock Mamba: Linear-time sequence modeling with selective state spaces.
\newblock \emph{arXiv preprint arXiv:2312.00752}, 2023.

\bibitem[Guu et~al.(2020)Guu, Lee, Tung, Pasupat, and Chang]{guu2020retrieval}
Kelvin Guu, Kenton Lee, Zora Tung, Panupong Pasupat, and Mingwei Chang.
\newblock Retrieval augmented language model pre-training.
\newblock In \emph{International conference on machine learning}, pp.\  3929--3938. PMLR, 2020.

\bibitem[Hsieh et~al.(2024)Hsieh, Sun, Kriman, Acharya, Rekesh, Jia, Zhang, and Ginsburg]{hsieh2024ruler}
Cheng-Ping Hsieh, Simeng Sun, Samuel Kriman, Shantanu Acharya, Dima Rekesh, Fei Jia, Yang Zhang, and Boris Ginsburg.
\newblock Ruler: What's the real context size of your long-context language models?
\newblock \emph{arXiv preprint arXiv:2404.06654}, 2024.

\bibitem[Huang et~al.(2025)Huang, Madala, Sidhu, Niu, Peng, Hockenmaier, and Zhang]{huang2025rag}
Jerry Huang, Siddarth Madala, Risham Sidhu, Cheng Niu, Hao Peng, Julia Hockenmaier, and Tong Zhang.
\newblock Rag-rl: Advancing retrieval-augmented generation via rl and curriculum learning.
\newblock \emph{arXiv preprint arXiv:2503.12759}, 2025.

\bibitem[Izacard \& Grave(2021)Izacard and Grave]{izacard2021leveraging}
Gautier Izacard and Edouard Grave.
\newblock Leveraging passage retrieval with generative models for open domain question answering.
\newblock In \emph{Proceedings of the 16th conference of the european chapter of the association for computational linguistics: main volume}, pp.\  874--880, 2021.

\bibitem[Izacard et~al.(2021)Izacard, Caron, Hosseini, Riedel, Bojanowski, Joulin, and Grave]{izacard2021unsupervised}
Gautier Izacard, Mathilde Caron, Lucas Hosseini, Sebastian Riedel, Piotr Bojanowski, Armand Joulin, and Edouard Grave.
\newblock Unsupervised dense information retrieval with contrastive learning.
\newblock \emph{arXiv preprint arXiv:2112.09118}, 2021.

\bibitem[Jiang et~al.(2024{\natexlab{a}})Jiang, Wu, Luo, Li, Lin, Yang, and Qiu]{jiang2024longllmlingua}
Huiqiang Jiang, Qianhui Wu, Xufang Luo, Dongsheng Li, Chin-Yew Lin, Yuqing Yang, and Lili Qiu.
\newblock Longllmlingua: Accelerating and enhancing llms in long context scenarios via prompt compression.
\newblock In \emph{Proceedings of the 62nd Annual Meeting of the Association for Computational Linguistics (Volume 1: Long Papers)}, pp.\  1658--1677, 2024{\natexlab{a}}.

\bibitem[Jiang et~al.(2024{\natexlab{b}})Jiang, Ma, and Chen]{jiang2024longrag}
Ziyan Jiang, Xueguang Ma, and Wenhu Chen.
\newblock Longrag: Enhancing retrieval-augmented generation with long-context llms.
\newblock \emph{arXiv preprint arXiv:2406.15319}, 2024{\natexlab{b}}.

\bibitem[Jin et~al.(2025)Jin, Zeng, Yue, Yoon, Arik, Wang, Zamani, and Han]{jin2025search}
Bowen Jin, Hansi Zeng, Zhenrui Yue, Jinsung Yoon, Sercan Arik, Dong Wang, Hamed Zamani, and Jiawei Han.
\newblock Search-r1: Training llms to reason and leverage search engines with reinforcement learning.
\newblock \emph{arXiv preprint arXiv:2503.09516}, 2025.

\bibitem[Karpukhin et~al.(2020)Karpukhin, Oguz, Min, Lewis, Wu, Edunov, Chen, and Yih]{karpukhin2020dense}
Vladimir Karpukhin, Barlas Oguz, Sewon Min, Patrick Lewis, Ledell Wu, Sergey Edunov, Danqi Chen, and Wen-tau Yih.
\newblock Dense passage retrieval for open-domain question answering.
\newblock In \emph{Proceedings of the 2020 conference on empirical methods in natural language processing (EMNLP)}, pp.\  6769--6781, 2020.

\bibitem[Kuratov et~al.(2024)Kuratov, Bulatov, Anokhin, Rodkin, Sorokin, Sorokin, and Burtsev]{kuratov2024babilong}
Yuri Kuratov, Aydar Bulatov, Petr Anokhin, Ivan Rodkin, Dmitry Sorokin, Artyom Sorokin, and Mikhail Burtsev.
\newblock Babilong: Testing the limits of llms with long context reasoning-in-a-haystack.
\newblock \emph{Advances in Neural Information Processing Systems}, 37:\penalty0 106519--106554, 2024.

\bibitem[Lewis et~al.(2020)Lewis, Perez, Piktus, Petroni, Karpukhin, Goyal, K{\"u}ttler, Lewis, Yih, Rockt{\"a}schel, et~al.]{lewis2020retrieval}
Patrick Lewis, Ethan Perez, Aleksandra Piktus, Fabio Petroni, Vladimir Karpukhin, Naman Goyal, Heinrich K{\"u}ttler, Mike Lewis, Wen-tau Yih, Tim Rockt{\"a}schel, et~al.
\newblock Retrieval-augmented generation for knowledge-intensive nlp tasks.
\newblock \emph{Advances in neural information processing systems}, 33:\penalty0 9459--9474, 2020.

\bibitem[Li et~al.(2024)Li, He, Guo, Bu, Bai, Liu, Liu, Qu, Li, Ouyang, et~al.]{li2024graphreader}
Shilong Li, Yancheng He, Hangyu Guo, Xingyuan Bu, Ge~Bai, Jie Liu, Jiaheng Liu, Xingwei Qu, Yangguang Li, Wanli Ouyang, et~al.
\newblock Graphreader: Building graph-based agent to enhance long-context abilities of large language models.
\newblock In \emph{Findings of the Association for Computational Linguistics: EMNLP 2024}, pp.\  12758--12786, 2024.

\bibitem[Mao et~al.(2021)Mao, He, Liu, Shen, Gao, Han, and Chen]{mao2021generation}
Yuning Mao, Pengcheng He, Xiaodong Liu, Yelong Shen, Jianfeng Gao, Jiawei Han, and Weizhu Chen.
\newblock Generation-augmented retrieval for open-domain question answering.
\newblock In \emph{Proceedings of the 59th Annual Meeting of the Association for Computational Linguistics and the 11th International Joint Conference on Natural Language Processing (Volume 1: Long Papers)}, pp.\  4089--4100, 2021.

\bibitem[Mnih et~al.(2015)Mnih, Kavukcuoglu, Silver, Rusu, Veness, Bellemare, Graves, Riedmiller, Fidjeland, Ostrovski, et~al.]{mnih2015human}
Volodymyr Mnih, Koray Kavukcuoglu, David Silver, Andrei~A Rusu, Joel Veness, Marc~G Bellemare, Alex Graves, Martin Riedmiller, Andreas~K Fidjeland, Georg Ostrovski, et~al.
\newblock Human-level control through deep reinforcement learning.
\newblock \emph{nature}, 518\penalty0 (7540):\penalty0 529--533, 2015.

\bibitem[Oord et~al.(2018)Oord, Li, and Vinyals]{oord2018representation}
Aaron van~den Oord, Yazhe Li, and Oriol Vinyals.
\newblock Representation learning with contrastive predictive coding.
\newblock \emph{arXiv preprint arXiv:1807.03748}, 2018.

\bibitem[Pereyra et~al.(2017)Pereyra, Tucker, Chorowski, Kaiser, and Hinton]{pereyra2017regularizing}
Gabriel Pereyra, George Tucker, Jan Chorowski, {\L}ukasz Kaiser, and Geoffrey Hinton.
\newblock Regularizing neural networks by penalizing confident output distributions.
\newblock \emph{arXiv preprint arXiv:1701.06548}, 2017.

\bibitem[Qi et~al.(2019)Qi, Lin, Mehr, Wang, and Manning]{qi2019answering}
Peng Qi, Xiaowen Lin, Leo Mehr, Zijian Wang, and Christopher~D Manning.
\newblock Answering complex open-domain questions through iterative query generation.
\newblock In \emph{Proceedings of the 2019 conference on empirical methods in natural language processing and the 9th international joint conference on natural language processing (EMNLP-IJCNLP)}, pp.\  2590--2602, 2019.

\bibitem[Rodkin et~al.(2024)Rodkin, Kuratov, Bulatov, and Burtsev]{rodkin2024associative}
Ivan Rodkin, Yuri Kuratov, Aydar Bulatov, and Mikhail Burtsev.
\newblock Associative recurrent memory transformer.
\newblock \emph{arXiv preprint arXiv:2407.04841}, 2024.

\bibitem[Shang et~al.(2025)Shang, Zhang, Wang, Zhang, Lopez, Yang, Chen, and Yang]{shang2025longrope2}
Ning Shang, Li~Lyna Zhang, Siyuan Wang, Gaokai Zhang, Gilsinia Lopez, Fan Yang, Weizhu Chen, and Mao Yang.
\newblock Longrope2: Near-lossless llm context window scaling.
\newblock \emph{arXiv preprint arXiv:2502.20082}, 2025.

\bibitem[Shao et~al.(2023)Shao, Gong, Shen, Huang, Duan, and Chen]{shao2023enhancing}
Zhihong Shao, Yeyun Gong, Yelong Shen, Minlie Huang, Nan Duan, and Weizhu Chen.
\newblock Enhancing retrieval-augmented large language models with iterative retrieval-generation synergy.
\newblock In \emph{Findings of the Association for Computational Linguistics: EMNLP 2023}, pp.\  9248--9274, 2023.

\bibitem[Singh et~al.(2025)Singh, Ehtesham, Kumar, Khoei, and Vasilakos]{singh2025agentic}
Aditi Singh, Abul Ehtesham, Saket Kumar, Tala~Talaei Khoei, and Athanasios~V Vasilakos.
\newblock Agentic retrieval-augmented generation: A survey on agentic rag.
\newblock \emph{arXiv preprint arXiv:2501.09136}, 2025.

\bibitem[Song et~al.(2025)Song, Jiang, Min, Chen, Chen, Zhao, Fang, and Wen]{song2025r1}
Huatong Song, Jinhao Jiang, Yingqian Min, Jie Chen, Zhipeng Chen, Wayne~Xin Zhao, Lei Fang, and Ji-Rong Wen.
\newblock R1-searcher: Incentivizing the search capability in llms via reinforcement learning.
\newblock \emph{arXiv preprint arXiv:2503.05592}, 2025.

\bibitem[Sorokin et~al.(2026)Sorokin, Buzun, Anokhin, Vedernikov, Anokhin, Burtsev, and Burnaev]{sorokin2026q}
Artyom Sorokin, Nazar Buzun, Aleksandr Anokhin, Egor Vedernikov, Petr Anokhin, Mikhail Burtsev, and Evgeny Burnaev.
\newblock Q-rag: Long context multi-step retrieval via value-based embedder training.
\newblock In \emph{International Conference on Learning Representations}, volume 2026, pp.\  32481--32506, 2026.

\bibitem[Su et~al.(2024)Su, Ahmed, Lu, Pan, Bo, and Liu]{su2024roformer}
Jianlin Su, Murtadha Ahmed, Yu~Lu, Shengfeng Pan, Wen Bo, and Yunfeng Liu.
\newblock Roformer: Enhanced transformer with rotary position embedding.
\newblock \emph{Neurocomputing}, 568:\penalty0 127063, 2024.

\bibitem[Szegedy et~al.(2016)Szegedy, Vanhoucke, Ioffe, Shlens, and Wojna]{szegedy2016rethinking}
Christian Szegedy, Vincent Vanhoucke, Sergey Ioffe, Jon Shlens, and Zbigniew Wojna.
\newblock Rethinking the inception architecture for computer vision.
\newblock In \emph{Proceedings of the IEEE conference on computer vision and pattern recognition}, pp.\  2818--2826, 2016.

\bibitem[Trivedi et~al.(2022)Trivedi, Balasubramanian, Khot, and Sabharwal]{trivedi2022musique}
Harsh Trivedi, Niranjan Balasubramanian, Tushar Khot, and Ashish Sabharwal.
\newblock musique: Multihop questions via single-hop question composition.
\newblock \emph{Transactions of the Association for Computational Linguistics}, 10:\penalty0 539--554, 2022.

\bibitem[Trivedi et~al.(2023)Trivedi, Balasubramanian, Khot, and Sabharwal]{trivedi2023interleaving}
Harsh Trivedi, Niranjan Balasubramanian, Tushar Khot, and Ashish Sabharwal.
\newblock Interleaving retrieval with chain-of-thought reasoning for knowledge-intensive multi-step questions.
\newblock In \emph{Proceedings of the 61st annual meeting of the association for computational linguistics (volume 1: long papers)}, pp.\  10014--10037, 2023.

\bibitem[Xiong et~al.(2020)Xiong, Li, Iyer, Du, Lewis, Wang, Mehdad, Yih, Riedel, Kiela, et~al.]{xiong2020answering}
Wenhan Xiong, Xiang~Lorraine Li, Srini Iyer, Jingfei Du, Patrick Lewis, William~Yang Wang, Yashar Mehdad, Wen-tau Yih, Sebastian Riedel, Douwe Kiela, et~al.
\newblock Answering complex open-domain questions with multi-hop dense retrieval.
\newblock \emph{arXiv preprint arXiv:2009.12756}, 2020.

\bibitem[Yang et~al.(2018)Yang, Qi, Zhang, Bengio, Cohen, Salakhutdinov, and Manning]{yang2018hotpotqa}
Zhilin Yang, Peng Qi, Saizheng Zhang, Yoshua Bengio, William Cohen, Ruslan Salakhutdinov, and Christopher~D Manning.
\newblock Hotpotqa: A dataset for diverse, explainable multi-hop question answering.
\newblock In \emph{Proceedings of the 2018 conference on empirical methods in natural language processing}, pp.\  2369--2380, 2018.

\bibitem[Yao et~al.(2022)Yao, Zhao, Yu, Du, Shafran, Narasimhan, and Cao]{yao2022react}
Shunyu Yao, Jeffrey Zhao, Dian Yu, Nan Du, Izhak Shafran, Karthik Narasimhan, and Yuan Cao.
\newblock React: Synergizing reasoning and acting in language models.
\newblock \emph{arXiv preprint arXiv:2210.03629}, 2022.

\bibitem[Zhang et~al.(2024)Zhang, Zhang, Zhang, Yong, and Huang]{zhang2024end}
Jiahao Zhang, Haiyang Zhang, Dongmei Zhang, Liu Yong, and Shen Huang.
\newblock End-to-end beam retrieval for multi-hop question answering.
\newblock In \emph{Proceedings of the 2024 Conference of the North American Chapter of the Association for Computational Linguistics: Human Language Technologies (Volume 1: Long Papers)}, pp.\  1718--1731, 2024.

\end{thebibliography}
\bibliographystyle{iclr2027_conference}

\newpage
\appendix
\section{\texorpdfstring{S$^2$LMR}{S2LMR} training procedure}
\label{app:ssr_alg}

Algorithm~\ref{alg:ssr} makes the training loop of Sec.~\ref{sec:ssr}
explicit for one batched update. Candidate chunks are encoded once per
example (line~5); at every hop the query state $s_t$ is rebuilt from the
gold prefix (line~8), the chunk-axis positions $p_j^{(t)}$ are recomputed
from the already-selected indices $\mathcal{S}_t$ (line~9), and candidate
embeddings are re-rotated accordingly (line~10). The per-hop InfoNCE loss
(line~12) instantiates Eq.~\ref{eq:hop_nce} on
$\mathcal{A}_t=\{1,\dots,N\}\setminus\mathcal{S}_t$, and the example loss
averages over the $K$ hops before the batch gradient step.

\begin{algorithm}[t]
\caption{S$^2$LMR: Supervised Sequential Long-Context Multi-Step Retriever (one training update)}
\label{alg:ssr}
\begin{algorithmic}[1]
\State \textbf{Hyperparameters:} batch size $B$, hops $K$, temperature $\tau$, RoPE step $\Delta$, learning rate $\eta$.
\State \textbf{Model:} shared bi-encoder $f_\theta$; chunk-axis $\mathrm{RoPE}(\cdot,p)$.
\State Sample batch $\bigl\{\bigl(q^{(b)},\{c_j^{(b)}\}_{j=1}^{N},(i_1^{\star,(b)},\dots,i_K^{\star,(b)})\bigr)\bigr\}_{b=1}^{B}$
\For{$b = 1,\dots,B$}
    \State $h_j \gets f_\theta(c_j^{(b)})$ for $j=1,\dots,N$ \Comment{encode candidates once}
    \State $\mathcal{S}_1 \gets \varnothing$
    \For{$t = 1,\dots,K$}
        \State build $s_t$ via Eq.~\ref{eq:state} with $c^{(k)} = c_{i_k^{\star,(b)}}^{(b)}$ for $k<t$ \Comment{teacher-forced state}
        \State compute $(p_j^{(t)})_{j=1}^{N}$ from $\mathcal{S}_t$ via Eq.~\ref{eq:pe}
        \State $\mathcal{A}_t \gets \{1,\dots,N\}\setminus \mathcal{S}_t$
        \State $\tilde{h}_j^{(t)} \gets \mathrm{RoPE}(h_j,\,p_j^{(t)})$ for all $j \in \mathcal{A}_t$
        \State $L_t^{(b)} \gets -\log\dfrac{\exp\!\bigl(\langle f_\theta(s_t),\tilde{h}_{i_t^{\star,(b)}}^{(t)}\rangle/\tau\bigr)}{\sum_{j\in\mathcal{A}_t}\exp\!\bigl(\langle f_\theta(s_t),\tilde{h}_j^{(t)}\rangle/\tau\bigr)}$ \Comment{Eq.~\ref{eq:hop_nce}}
        \State $\mathcal{S}_{t+1} \gets \mathcal{S}_t \cup \{i_t^{\star,(b)}\}$
    \EndFor
    \State $L^{(b)} \gets \tfrac{1}{K}\sum_{t=1}^{K} L_t^{(b)}$
\EndFor
\State $\mathcal{L} \gets \tfrac{1}{B}\sum_{b=1}^{B} L^{(b)}$;\ \ $\theta \gets \theta - \eta\,\nabla_\theta \mathcal{L}$
\end{algorithmic}
\end{algorithm}

\section{Connection Between Value-Based RL and Supervised Retrieval}
\label{app:rl-supervised}

This section formalizes the connection between the value-based RL formulation
of Q-RAG and direct supervised learning for multi-step retrieval. We show that
under a particular limiting regime, namely $T=K$, $\lambda=1$, and
$\alpha\rightarrow0$, the Q-RAG learning signal reduces to binary supervision
over sampled state--action pairs.
This does not imply that the Q-RAG and S$^2$LMR objectives are mathematically
identical, rather, it shows that the distinctive components of value-based RL
become unnecessary for constructing the learning signal in this regime.

\subsection{Problem Setup}

Consider a multi-step retrieval problem with $K$ ground-truth supporting facts.
At retrieval step $t$, the state consists of the query and the supporting facts
retrieved at previous steps. S$^2$LMR constructs the state $s_t$ by teacher
forcing (Eq.~\ref{eq:state} with $c^{(k)}=c_{i_k^\star}$ for $k<t$). For
brevity, let
\begin{align*}
\tilde h_j^{(t)} \;=\; \operatorname{RoPE}\!\big(f_\theta(c_j),\,p_j^{(t)}\big)
\end{align*}
denote the rotated candidate embedding at hop $t$, so that
$\operatorname{score}(s_t,c_j)=\langle f_\theta(s_t),\,\tilde h_j^{(t)}\rangle$
(Eq.~\ref{eq:score}). The per-hop InfoNCE objective from Eq.~\ref{eq:hop_nce}
can then be written as
\begin{equation}
\mathcal{L}^{\mathrm{S^2LMR}}_t
=
-\log
\frac{\exp\!\big(\langle f_\theta(s_t),\,\tilde h_{i_t^\star}^{(t)}\rangle/\tau\big)}
     {\sum_{j\in\mathcal{A}_t}\exp\!\big(\langle f_\theta(s_t),\,\tilde h_j^{(t)}\rangle/\tau\big)},
\label{eq:app_ssr}
\end{equation}
where $\tau$ is the contrastive temperature and
$\mathcal{A}_t=\mathcal{C}\setminus\{i_1^\star,\ldots,i_{t-1}^\star\}$ is the
remaining candidate set at hop $t$.

Q-RAG instead defines a value function over state--action pairs,
\begin{align*}
Q(s_t,c_j)
\;=\;
\langle \phi(s_t),\,\psi(c_j)\rangle,
\end{align*}
and uses a Boltzmann policy
\begin{equation}
\pi_\alpha(c_j\mid s_t)
\;=\;
\frac{\exp(Q(s_t,c_j)/\alpha)}
     {\sum_{j'\in\mathcal{A}_t}\exp(Q(s_t,c_{j'})/\alpha)},
\label{eq:app_policy}
\end{equation}
where $\alpha$ controls the degree of stochastic exploration. The value
function is trained by minimizing the squared error to a $\lambda$-return:
\begin{align*}
\mathcal{L}^{\mathrm{Q\text{-}RAG}}
\;=\;
\mathbb{E}\!\left[\big(Q(s_t,c_{i_t})-G_t^\lambda\big)^{\!2}\right].
\end{align*}

For the multi-hop QA setting considered here, the terminal reward is
constructed from the ground-truth supporting-fact indices:
\begin{align*}
r_T \;=\; \mathbbm{1}\!\left[\{i_1^\star,\ldots,i_K^\star\} \subseteq \{i_1,\ldots,i_T\}\right].
\end{align*}
Thus, a trajectory receives positive reward when the retrieved set covers all
annotated supporting facts.

\subsection{Effect of the Retrieval Horizon}

We first consider the limiting case in which the retrieval horizon equals the
number of supporting facts:
\begin{align*}
T \;=\; K.
\end{align*}

Assume that additional retrieval beyond the $K$ required supporting facts is
penalized, as in the setting considered in our analysis. Since a successful
trajectory must contain all $K$ supporting facts and contains exactly $K$
retrieval steps, a successful trajectory cannot contain an additional
non-supporting passage. Therefore,
\begin{align*}
r_T
\;=\;
\begin{cases}
1, & \{i_1,\ldots,i_K\}=\{i_1^\star,\ldots,i_K^\star\}, \\
0, & \text{otherwise}.
\end{cases}
\end{align*}

This condition removes the ambiguity caused by extra retrieved passages. In
the general set-based reward formulation, a trajectory may receive reward $1$
even if it contains irrelevant passages in addition to all supporting facts.
When $T=K$, this ambiguity disappears because there is no additional retrieval
step available after covering the $K$ supporting facts.

\subsection{Effect of the \texorpdfstring{$\lambda$}{lambda}-Return}

Next, consider the case
\begin{align*}
\lambda \;=\; 1.
\end{align*}

With only a terminal reward, the $\lambda$-return reduces to the Monte Carlo
return:
\begin{align*}
G_t^\lambda \;\overset{\lambda=1}{=}\; G_t^{\mathrm{MC}} \;=\; r_T,
\end{align*}
up to the discounting convention used by the implementation. In the
undiscounted setting considered here, this gives
$G_t^\lambda \in \{0,1\}$.

Consequently, the Q-RAG objective becomes
\begin{align*}
\mathcal{L}^{\mathrm{Q\text{-}RAG}}
\;=\;
\mathbb{E}\!\left[\big(Q(s_t,c_{i_t})-y_t\big)^{\!2}\right],
\qquad y_t=r_T\in\{0,1\}.
\end{align*}

Thus, the target no longer depends on a bootstrapped estimate of a future
value. A state--action pair on a successful trajectory receives target $1$,
whereas a state--action pair on a failed trajectory receives target $0$. This
removes the bootstrapping component that distinguishes the general
temporal-difference formulation.

\subsection{Deterministic Policy Limit}

Finally, consider the limit
\begin{align*}
\alpha \;\rightarrow\; 0.
\end{align*}

For the Boltzmann policy in Eq.~\ref{eq:app_policy}, the probability mass
concentrates on actions with the maximum value:
\begin{align*}
\lim_{\alpha\rightarrow 0}\pi_\alpha(c_j\mid s_t)
\;=\;
\begin{cases}
1, & c_j\in\arg\max_{c_{j'}}Q(s_t,c_{j'}),\\
0, & \text{otherwise},
\end{cases}
\end{align*}
with an arbitrary tie-breaking rule when multiple actions attain the same
maximum. Hence, in this limit, sampling becomes equivalent to greedy action
selection:
\begin{align*}
i_t \;=\; \arg\max_{j\in\mathcal{A}_t}Q(s_t,c_j).
\end{align*}

The stochastic exploration component of the policy is therefore removed.
Importantly, this limit alone does not guarantee a successful trajectory;
rather, once the policy has converged to a successful deterministic policy,
the resulting rollouts repeatedly expose state--action pairs that lead to
successful retrieval.

\subsection{Resulting Learning Signal}

Combining $T=K$, $\lambda=1$, and $\alpha\rightarrow0$, the terminal outcome
provides a binary target for each sampled state--action pair:
\begin{align*}
y_t \;=\;
\begin{cases}
1, & (s_t,c_{i_t}) \text{ on a successful trajectory}, \\
0, & (s_t,c_{i_t}) \text{ on a failed trajectory}.
\end{cases}
\end{align*}

The Q-RAG objective consequently takes the form
\begin{align*}
\mathcal{L}^{\mathrm{Q\text{-}RAG}}
\;=\;
\mathbb{E}\!\left[\big(Q(s_t,c_{i_t})-y_t\big)^{\!2}\right],
\qquad y_t\in\{0,1\}.
\end{align*}

The corresponding gradient with respect to the parameters of $Q$ is
\begin{align*}
\nabla \mathcal{L}^{\mathrm{Q\text{-}RAG}}
\;=\;
2\big(Q(s_t,c_{i_t})-y_t\big)\,\nabla Q(s_t,c_{i_t}).
\end{align*}

Therefore, successful state--action pairs are pushed toward value $1$, while
state--action pairs observed on failed trajectories are pushed toward value
$0$. In this limiting regime, the Q-RAG learning signal can thus be viewed as
a binary supervised regression objective over sampled trajectory actions.

\subsection{Relation to \texorpdfstring{S$^2$LMR}{S2LMR}}

S$^2$LMR instead uses the contrastive objective in Eq.~\ref{eq:app_ssr}. Its
gradient with respect to the query-state embedding is
\begin{align*}
\nabla_{f_\theta(s_t)}\mathcal{L}^{\mathrm{S^2LMR}}_t
\;=\;
-\frac{1}{\tau}\Bigg(
\tilde h_{i_t^\star}^{(t)}
\;-\;
\sum_{j\in\mathcal{A}_t}
p_\tau(j\mid s_t)\,\tilde h_j^{(t)}
\Bigg),
\end{align*}
where
\begin{align*}
p_\tau(j\mid s_t)
\;=\;
\frac{\exp(\langle f_\theta(s_t),\,\tilde h_j^{(t)}\rangle/\tau)}
     {\sum_{j'\in\mathcal{A}_t}\exp(\langle f_\theta(s_t),\,\tilde h_{j'}^{(t)}\rangle/\tau)}.
\end{align*}

Thus, S$^2$LMR increases the score of the ground-truth supporting passage
while decreasing the scores of competing candidates, with the magnitude of the
negative signal determined by their current softmax probabilities.

The two objectives are therefore not mathematically identical. Their
difference is summarized in Table~\ref{tab:app_rl_ssr}.

\begin{table}[t]
\centering
\small
\caption{
Comparison between S$^2$LMR and the limiting form of Q-RAG.
}
\label{tab:app_rl_ssr}
\begin{tabular}{lcc}
\toprule
Property & S$^2$LMR & Q-RAG limiting case \\
\midrule
Learning signal
& Fixed per-hop supervision
& Sampled trajectory outcome \\
Objective
& InfoNCE
& Binary MSE \\
Positive signal
& Ground-truth supporting passage
& Action on successful trajectory \\
Negative signal
& All in-context candidates
& Sampled actions on failed trajectories \\
Bootstrapping
& None
& None ($\lambda=1$) \\
Exploration
& None
& None in the limit ($\alpha\rightarrow0$) \\
State construction
& Teacher forcing
& Policy-generated trajectory \\
\bottomrule
\end{tabular}
\end{table}

The key connection is therefore not an exact equivalence between the two loss
functions. Rather, under the limiting regime considered above, the Q-RAG
formulation loses the main components that distinguish value-based RL from
direct supervised learning: extra-step credit assignment, bootstrapping, and
stochastic exploration. What remains is a supervised learning signal that
encourages relevant actions to receive higher scores than sampled irrelevant
actions. S$^2$LMR directly provides this signal through fixed per-hop
supervision and contrasts each positive passage against all in-context
candidates.

This analysis motivates our empirical comparison: when supporting-fact
supervision is available, value-based RL introduces additional optimization
machinery beyond the direct supervised signal, while it remains an empirical
question whether that additional machinery yields better retrieval
performance.

\section{Additional Results}
\subsection{Empirical evidence for Issue 1 (distractor-distribution mismatch)}
\label{app:m1_evidence}

To isolate what MDR-style in-batch training actually optimizes, we
compare the encoder's hop-1 discrimination on the two negative
distributions that matter---the \emph{training pool} (other in-batch
queries' gold chunks) and the \emph{inference pool} (same-document
distractors)---across three checkpoints: (i) a zero-shot Contriever
initialization, (ii) an MDR-style reproduction trained on in-batch
negatives, and (iii) the S$^2$LMR checkpoint trained on same-context
negatives. 
All three share the same encoder, tokenizer, and RoPE positional encoding. 
Query states are teacher-forced with the true hop-0 answer so measurements isolate per-hop discrimination.

\begin{table}[t]
\centering\small
\caption{Hop-1 median cosine similarities on 128 BabiLong QA3 samples
(RoPE-rotated). 
$\text{margin}_{\text{sc}} = \text{sim(gold)} - \text{sim(sc-neg)}$;
$\text{margin}_{\text{ib}} = \text{sim(gold)} - \text{sim(ib-neg)}$. MDR training widens the
in-batch margin more than the same-context margin (its own objective
converges on the wrong distribution). S$^2$LMR training widens the
same-context margin most, matching what inference requires. All three
checkpoints use identical encoder/tokenizer/RoPE.}
\begin{tabular}{lccccc}
\toprule
Checkpoint & sim(gold) & sim(sc-neg) & sim(ib-neg) &
margin$_{\text{sc}}$ & margin$_{\text{ib}}$ \\
\midrule
Contriever (zero-shot)          & 0.29 & 0.23 & 0.23 & +0.06 & +0.06 \\
MDR-repro (in-batch trained)    & 0.45 & 0.36 & 0.34 & +0.09 & \textbf{+0.11} \\
S$^2$LMR (same-context trained)      & 0.67 & 0.46 & 0.54 & \textbf{+0.21} & +0.13 \\
\bottomrule
\end{tabular}
\label{tab:m1_evidence}
\end{table}

\begin{figure}
    \centering
    \includegraphics[width=\linewidth]{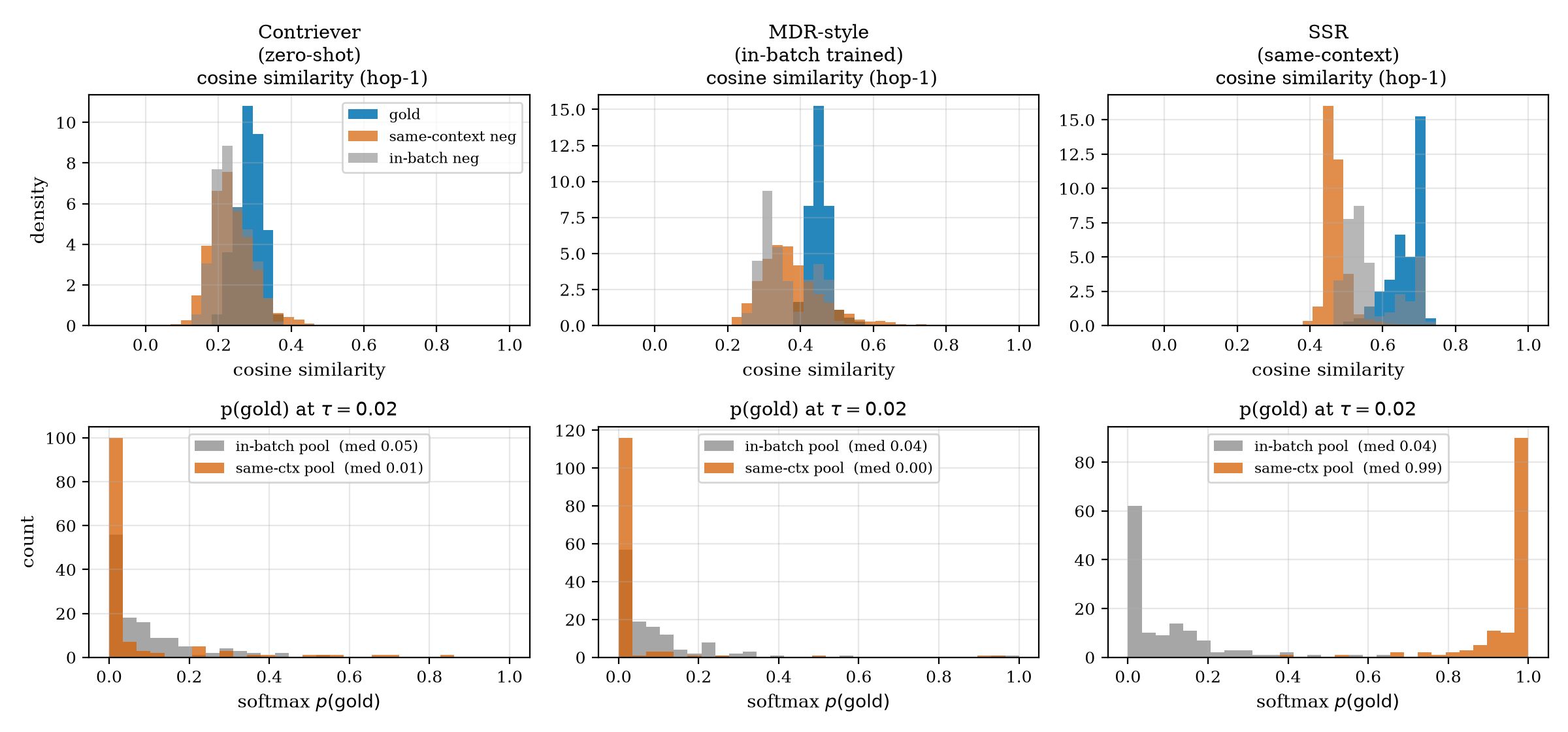}
    \caption{Distractor-distribution mismatch on 128 BabiLong QA3 samples. \textbf{Top row:} distributions of cosine similarity between the query state and (i) the gold chunk, (ii) 31 same-document distractors, (iii) 31 in-batch negatives. \textbf{Bottom row:} distributions of softmax $p(\mathrm{gold})$ at $\tau{=}0.02$ under the in-batch pool (MDR training pool, 32 candidates) and the same-context pool (inference pool, 32 candidates). Columns: Contriever (zero-shot), MDR-style checkpoint trained with in-batch negatives, and S$^2$LMR trained with same-context negatives.
    }
    \label{fig:m1_distributions}
\end{figure}

The pattern is symmetric at initialization ($\text{margin}_{\text{sc}} {=} \text{margin}_{\text{ib}} {=} 0.06$)
and diverges opposite to inference needs under MDR training ($\text{margin}_{\text{ib}} {>} \text{margin}_{\text{sc}}$).
MDR's gradient signal itself is strong (median $p(\text{gold})$ on its in-batch pool is 0.04, implying gradient-magnitude proxy $\approx 0.96$), so the failure is not gradient collapse but distributional misalignment---the encoder converges to a boundary that separates the wrong distribution. 
Consequently the MDR checkpoint ranks the hop-1 gold above the 99 same-document distractors on 0 of 128 samples, whereas S$^2$LMR does so on all 128 (Figure~\ref{fig:m1_distributions}).

\subsection{Teacher forcing or not: supervision noise vs. exposure bias}
\label{app:exposure_bias}
BR-style beam-expanded supervision was originally motivated by an exposure-bias concern: training only on gold prefixes leaves the model unfamiliar with the noisier trajectories it will produce at inference. S$^2$LMR accepts this trade-off because the alternative---training on noisy predicted prefixes---is empirically more harmful in the long-context regime.
Fig.~\ref{fig:m2_failure} and Fig.~\ref{fig:method_effect}b show that as context length grows, the beam-expanded performance degrades noticeably faster than the teacher-forced curve, while at the training length the two are within $0.04$ Fact EM (Figure~\ref{fig:m2_failure}, Issue 2).
The exposure gap that teacher forcing incurs is small enough to be dominated by per-hop InfoNCE against the full same-context pool, which keeps the gold-vs-distractor margin large enough that small prefix perturbations at inference rarely flip the argmax.

\subsection{Positional bottleneck in chunk-position-agnostic retrievers}
\label{app:pe_bottleneck}

This appendix expands the analysis in Sec.~\ref{sec:motivation_ssr} of the positional-encoding bottleneck in existing supervised multi-hop retrievers.
We (i) enumerate the architectural facts that make MDR and BR chunk-position-agnostic, (ii) quantify the positional structure of BabiLong QA3 that these architectures cannot exploit, (iii) measure the resulting per-hop retrieval margin, and (iv) show the failure mode is stable across all three hops.

\subsubsection{Architecture: token-axis only positional encoding}
\label{app:pe_arch}

\paragraph{MDR~\citep{xiong2020answering}.}
MDR uses a shared RoBERTa-base bi-encoder to embed candidates and queries independently. 
Each candidate chunk $c_j$ is embedded as $E_a(c_j) = \operatorname{RoBERTa}(c_j)$. 
Only tokens within $c_j$ receive positional information via RoBERTa's absolute learned positional embeddings (maximum position 514). 
The resulting embedding depends solely on the text of $c_j$ and is invariant to $c_j$'s document-level position. 
At the query side, MDR concatenates the question with previously retrieved passages and re-encodes. 
The maximum sequence length is again 514 tokens, and no chunk-level position is exposed.

\paragraph{BR~\citep{zhang2024end}.}
BR uses a DeBERTa cross-encoder that scores candidates in flat concatenation $[\text{CLS}; Q; \hat{p}_1; \ldots; \hat{p}_{t-1}; c_j; \text{SEP}]$ (maximum 512 tokens). 
DeBERTa's disentangled relative positional encoding operates on token-token offsets within this flat sequence. 
BR's Appendix~E \emph{explicitly shuffles the order of concatenated passages during training} to enforce robustness to permutations. 
This is architecturally equivalent to training the encoder to be order-invariant across chunks: any positional distinction between the concatenated passages that could have arisen from token positions in the flat sequence is deliberately regularized away.

\paragraph{Common consequence.}
In both architectures, the retrieval decision at hop $t$ depends only on the \emph{textual} content of $c_j$ and the (question, previous picks) tuple, never on where $c_j$ sits in the source document. We call this property \emph{chunk-position-agnostic}.

\subsubsection{Task-side positional structure in BabiLong QA3}
\label{app:pe_task}

BabiLong~\citep{kuratov2024babilong} embeds bAbI supporting facts as a temporally-ordered subsequence within a long stream of PG-19 noise sentences. 
QA3 (``three-supporting-facts'') requires retrieving three temporally consecutive facts about a person's location history. 
Because the story is ordered in time and the question refers to a temporal anchor (``\emph{Where was X before Y?}''), each gold fact appears in a specific positional relation to the others.

\paragraph{Case study.}
Consider a QA3 sample from the test set with question \emph{``Where was the apple before the bathroom?''} and answer \emph{``office''}. 
The three supporting facts, ordered as the model must retrieve them under fixed-order supervision, are:
\begin{enumerate}
  \item (chunk 48): \emph{``John discarded the apple there.''}
  \item (chunk 31): \emph{``John journeyed to the bathroom.''}
  \item (chunk 27): \emph{``John went to the office.''}
\end{enumerate}
The doc positions are strictly decreasing: 48 $\to$ 31 $\to$ 27. 
This is determined by the semantics of the question --- the temporal marker ``\emph{before}'' places each subsequent gold \emph{earlier} in the story than the previously picked one. 
A question with ``\emph{after}'' would invert the direction.
The direction is therefore \emph{semantically conditional}: which side of the picked chunk to search depends on the question, and cannot be reduced to a fixed rule.

\paragraph{Statistical structure.}
Table~\ref{tab:offset_stats} reports the distribution of the normalized offset $|\delta_t|/N$, where $\delta_t = \text{sf}_t - \text{sf}_{t-1}$, across 500 QA3 samples at each $N \in \{100, 800, 6400\}$.
Two properties stand out:
\begin{enumerate}
\item \textbf{Concentration.} The median $|\delta_t|/N$ is $0.05$--$0.06$ (vs.\ $0.33$ under uniform sampling), and $70\%$ of samples have $|\delta_t|/N \le 0.10$.
The IQR is $0.07$--$0.09$ (vs.\ $0.50$ uniform). 
A retriever that could restrict candidates to $|\text{position} - \text{last pick}| \le 0.10N$ would retain 70\% of golds while shrinking the candidate pool $10\times$.
\item \textbf{Scale invariance.} The distribution normalized by $N$ is       essentially identical at $N=100$, $N=800$, and $N=6400$: the positional structure is a property of the task, not of the context size.
\end{enumerate}

\begin{table}[t]
\centering
\small
\setlength{\tabcolsep}{6pt}
\caption{
\textbf{Concentration of next-gold document-position offsets in BabiLong QA3}, computed on 500 samples per $N$. 
The last column reports the fraction of samples with $|\delta_t|/N \le 0.10$ (range across the two hop transitions).
The distribution is scale-invariant in $N$ and far tighter than the uniform baseline, giving any position-aware retriever a substantial candidate-pool reduction.
}
\begin{tabular}{lccccccc}
\toprule
& \multicolumn{3}{c}{$|\delta_2|/N$}
& \multicolumn{3}{c}{$|\delta_3|/N$}
& $P(|\delta_t|/N$ \\
\cmidrule(lr){2-4}\cmidrule(lr){5-7}
$N$ & median & IQR & 90th pct & median & IQR & 90th pct & $\le 0.10)$ \\
\midrule
Uniform & 0.333 & 0.500 & 0.900 & 0.333 & 0.500 & 0.900 & 0.190 \\
\midrule
100     & 0.060 & 0.090 & 0.200 & 0.050 & 0.070 & 0.210 & 0.70--0.79 \\
800     & 0.056 & 0.092 & 0.218 & 0.048 & 0.089 & 0.182 & 0.69--0.73 \\
6400    & 0.055 & 0.088 & 0.214 & 0.049 & 0.087 & 0.193 & 0.71--0.73 \\
\bottomrule
\end{tabular}
\label{tab:offset_stats}
\end{table}

The case study and statistics together establish a claim we will use repeatedly: \emph{document position is a strong, semantically-conditional retrieval signal in BabiLong QA3}. 
A chunk-position-agnostic encoder cannot access this signal.

\subsubsection{Semantic-only ranking fails at every hop}
\label{app:pe_hop1}

We now measure the retrieval margin achievable by a fine-tuned bi-encoder that mirrors MDR's chunk-position-agnostic architecture.

\paragraph{Proxy encoder.}
We use the Contriever~\citep{izacard2021unsupervised} bi-encoder architecture, fine-tuned on BabiLong QA3 ($N=100$) for 2000 optimizer steps with per-hop InfoNCE loss, label smoothing $0.05$, and CCR regularization $\lambda = 0.02$, with chunk-axis positional encoding disabled. 
Every element of the training setup matches our other BabiLong retrievers except the absence of chunk-axis positional information.
The resulting encoder is architecturally chunk-position-agnostic while task-specifically fine-tuned.

\paragraph{Metric.}
For each sample and each hop $t$, we encode the query context (question concatenated with oracle previous gold picks $\text{sf}_1, \ldots, \text{sf}_{t-1}$), score all remaining chunks by inner product with the query embedding, and record
\begin{align*}
  \text{gold\_max\_cos}_t   &= \max_{j \in \text{gold}} \cos(q_t, c_j), \\
  \text{best\_noise\_cos}_t &= \max_{j \notin \text{gold}} \cos(q_t, c_j), \\
  \text{margin}_t           &= \text{gold\_max\_cos}_t - \text{best\_noise\_cos}_t.
\end{align*}
$P(\text{margin}_t \le 0)$ is the fraction of samples on which the top-scoring chunk (which the retriever picks by argmax) is a distractor rather than a gold.

\paragraph{Scale invariance.}
Table~\ref{tab:hop1_margin_sweep} sweeps $N$ from 10 to 6400 at hop 1. 
Both $\text{gold\_max\_cos}$ and $\text{best\_noise\_cos}$ are stable (0.962 vs.\ 0.963, respectively), the median margin sits at $-0.001$, and $P(\text{margin} \le 0)$ ranges only from $0.655$ to $0.670$.
Fine-tuning has successfully pushed both gold and distractors to high cosine ($\approx 0.96$), but has not opened any margin between them. 
And this failure is independent of the candidate pool size, ruling out ``noise-tail-grows-with-$N$'' as the explanation. The bottleneck is the encoder's inability to semantically separate a gold bAbI fact from a same-domain distractor.

\begin{table}[t]
\centering
\small
\setlength{\tabcolsep}{5pt}
\caption{
\textbf{Hop-1 semantic margin of a fine-tuned chunk-position-agnostic bi-encoder is scale-invariant in $N$.} 
Aggregated over 200 samples per $N$. 
The margin is $\approx 0$ regardless of context size.
}
\begin{tabular}{rcccccc}
\toprule
$N$ & 10 & 20 & 50 & 100 & 800 & 6400 \\
\midrule
$\text{gold\_max\_cos}$ (median)   & 0.962 & 0.962 & 0.962 & 0.962 & 0.962 & 0.962 \\
$\text{best\_noise\_cos}$ (median) & 0.963 & 0.963 & 0.963 & 0.963 & 0.963 & 0.963 \\
margin (median)                    & $-$0.001 & $-$0.001 & $-$0.001 & $-$0.001 & $-$0.001 & $-$0.001 \\
$P(\text{margin} \le 0)$           & 0.670 & 0.670 & 0.670 & 0.670 & 0.670 & 0.655 \\
\bottomrule
\end{tabular}
\label{tab:hop1_margin_sweep}
\end{table}

\paragraph{Why fine-tuning cannot separate gold from distractors.}
BabiLong noise chunks are sampled from PG-19 book text but interleaved with bAbI-style distractor sentences that share the same syntactic template as the gold facts (subject-verb-location: ``\emph{Mary moved to the office.}'').
Under InfoNCE fine-tuning, the encoder learns to give high cosine to any (question, bAbI-template sentence about the queried entity) pair. 
Because the distractors are indistinguishable from golds under a purely semantic similarity criterion --- both mention the queried entity, both involve a location and an action --- the fine-tune procedure pushes them to the same region of the embedding space. 
Positional information is the only signal that separates them, and this signal is absent by construction.

\subsubsection{Per-hop failure persists across all three hops}
\label{app:pe_perhop}

Table~\ref{tab:perhop_margin} extends the measurement to hops 2 and 3, with oracle previous gold picks fed as prev-context. 
Two observations sharpen the diagnosis:

\begin{itemize}
\item \textbf{The failure mode persists at every hop.}
      $P(\text{margin} \le 0)$ is $0.66$ at hop 1, $0.53$ at hop 2, and       $0.80$ at hop 3 (all at $N=100$). 
      The chunk-position-agnostic architecture fails to distinguish gold from distractor at every step of the multi-hop chain, not merely at hop 1.
\item \textbf{Hop 3 is the hardest.}
      Hop 2 is slightly easier because the single gold prev-context injects task-specific semantic content that the encoder can exploit ($P(\text{margin} \le 0) = 0.53$ vs.\ 0.66 at hop 1). Hop 3 is the hardest ($P = 0.80$): with two prev-context chunks the query context is longer and more diffuse, and the encoder's discriminative capacity is exceeded.
\end{itemize}

\rev{\paragraph{Joint K=3 retrieval success.}
Under oracle previous picks (200 QA3 samples), the chunk-position-agnostic encoder scores the gold chunk above all non-gold chunks at every hop in only 8.5\% of samples, versus 98--99\% with chunk-axis PE.}

\begin{table}[t]
\centering
\small
\setlength{\tabcolsep}{6pt}
\caption{
\textbf{Per-hop $P(\text{margin} \le 0)$ under oracle previous picks.}
The failure rate is stable at every hop across $N$. Hop 3 is the hardest, hop 2 marginally easier due to the injected gold-prev-context.
}
\begin{tabular}{lccc}
\toprule
$N$ & Hop 1 & Hop 2 & Hop 3 \\
\midrule
100  & 0.670 & 0.530 & 0.795 \\
800  & 0.670 & 0.520 & 0.795 \\
6400 & 0.660 & 0.485 & 0.795 \\
\bottomrule
\end{tabular}
\label{tab:perhop_margin}
\end{table}

\subsection{Q-RAG baseline: sources and evaluation protocols}
\label{app:qrag_protocol}
\rev{Q-RAG is our main baseline. Because its published numbers, released code, and released
checkpoints do not always agree, this section states where each Q-RAG number comes from and
how evaluation protocols differ. In all main comparisons, S$^2$LMR and Q-RAG use the same data,
evaluation script, and reader.}

\rev{\paragraph{Sources.} We evaluate the checkpoints released by ~\citep{sorokin2026q} on RULER QA,
HotpotQA and MuSiQue, and retrain Q-RAG with the configuration in the official repository on BabiLong,
HotpotQA and MuSiQue. The repository configuration differs from that of the released
checkpoints, which likely explains why retraining scores slightly lower (HotpotQA Fact F1/EM 0.92/0.87
vs.\ 0.93/0.89; MuSiQue 0.80/0.69 vs.\ 0.81/0.72 under Q-RAG's protocol); we therefore report both.
The training-dynamics study (Section~\ref{sec:exp:stability:dynamics}, Figures~\ref{fig:hparam_range}--\ref{fig:stability}) requires training runs and uses our own Q-RAG
runs with the paper's best reported setting. Figure~\ref{fig:babilong} follows Q-RAG's evaluation protocol, a budget of
$T{=}6$ retrieval steps credited whenever all gold facts are retrieved, whereas Figure~\ref{fig:stability:peak} evaluates both
methods with $T{=}K{=}3$ steps that must all be correct.}

\rev{\paragraph{MuSiQue.}
Q-RAG's evaluation allows up to four retrieval steps and stops an episode once all gold facts have
been retrieved, a stopping rule that uses gold labels. S$^2$LMR retrieves exactly $K=|\text{gold}|$
passages. Table~\ref{tab:qrag_musique_protocol} reports both methods under both protocols; S$^2$LMR
is ahead under each.}

\begin{table}[t]
\centering
\caption{\rev{MuSiQue (full dev) under two retrieval protocols. Q-RAG: released checkpoint.
S$^2$LMR: Fact is the mean of two seeds; Ans is seed 42. Answers by QwQ-32B.}}
\label{tab:qrag_musique_protocol}
\begin{tabular}{@{}llcccc@{}}
\toprule
Protocol & Method & Fact F1 & Fact EM & Ans F1 & Ans EM \\
\midrule
$K=|\text{gold}|$ & Q-RAG & 0.78 & 0.57 & 0.55 & 0.41 \\
                  & S$^2$LMR & \textbf{0.83} & \textbf{0.67} & \textbf{0.56} & \textbf{0.42} \\
\midrule
$\le$4 steps, stop when all gold found & Q-RAG & 0.81 & 0.72 & 0.60 & 0.45 \\
                                       & S$^2$LMR & \textbf{0.86} & \textbf{0.77} & \textbf{0.62} & \textbf{0.47} \\
\bottomrule
\end{tabular}
\end{table}

\rev{\paragraph{RULER SH/MH QA.}
Q-RAG's paper (Appendix~G) uses $T=4$ retrieval steps, whereas its released evaluation commands use
$T=1$ for SH and $T=3$ for MH. Table~2 uses the released setting. Table~\ref{tab:qrag_ruler_protocol}
reports both settings with the released checkpoint, alongside the published numbers. The $T=4$ setting
reproduces the published SH scores within two points. Under either setting, S$^2$LMR matches or
exceeds Q-RAG on MH at every length; on SH it is better at 4K and behind at longer contexts.}

\begin{table}[t]
\centering
\caption{\rev{RULER QA (EM) at 4K/16K/32K/128K/1M. Both methods are trained on HotpotQA+MuSiQue with
gte-multilingual-base and use the Qwen3-4B reader (no thinking), as in Q-RAG's Appendix~G.}}
\label{tab:qrag_ruler_protocol}
\begin{tabular}{@{}llcc@{}}
\toprule
Setting & Method & SH & MH \\
\midrule
Code ($T=1$/$3$) & Q-RAG    & 55/53/52/50/50 & 64/60/63/61/56 \\
                 & S$^2$LMR & 56/51/50/47/46 & 64/62/63/62/59 \\
\midrule
Paper ($T=4$)    & Q-RAG    & 61/59/59/57/53 & 62/60/58/60/57 \\
                 & S$^2$LMR & 63/57/57/55/45 & 65/61/63/63/61 \\
\midrule
Published        & Q-RAG    & 62/59/59/55/52 & 67/64/65/65/61 \\
\bottomrule
\end{tabular}
\end{table}

\subsection{Hyperparameter Ablation}
\label{sec:ablation}
Beyond robustness to training hyperparameters (Section~\ref{sec:exp:stability:hp}), we evaluate
S$^2$LMR's sensitivity to its three method-specific hyperparameters: the InfoNCE temperature $\tau$,
the confidence-regularizer weight $\lambda_{\mathrm{CCR}}$, and the label-smoothing coefficient
$\alpha_{\mathrm{LS}}$.

\rev{\paragraph{Regularized objective.}
Let $P_\theta(j \mid s_t)$ denote the softmax over $\mathcal{A}_t$ in Eq.~4, so that
$\mathcal{L}_t = -\log P_\theta(i^\star_t \mid s_t)$. In practice, each per-hop term is
\begin{equation}
\tilde{\mathcal{L}}_t = (1-\alpha_{\mathrm{LS}})\,\mathcal{L}_t
- \frac{\alpha_{\mathrm{LS}}}{|\mathcal{A}_t|}\sum_{j \in \mathcal{A}_t} \log P_\theta(j \mid s_t)
- \lambda_{\mathrm{CCR}}\, H\big(P_\theta(\cdot \mid s_t)\big),
\end{equation}
i.e., cross-entropy with label smoothing~\citep{szegedy2016rethinking} plus a confidence
regularizer (CCR) that rewards a higher-entropy candidate distribution, the confidence penalty
of \citet{pereyra2017regularizing}. Defaults are $\alpha_{\mathrm{LS}}=0.05$ and
$\lambda_{\mathrm{CCR}}=0.02$ unless Table~\ref{tab:hparams} states otherwise.}

Across a $10\times$ range of each hyperparameter, S$^2$LMR's Fact EM stays within a narrow band---$>0.94$ on QA2, $>0.97$ on QA3, and within a $\sim 0.04$ band around $0.88$ on HotpotQA---degrading only at extreme values ($\tau = 0.2$ or $\alpha_{\mathrm{LS}} \geq 0.05$). 
The three benchmarks share overlapping optimal regions across all three hyperparameters, so a single default setting transfers across both long-context and open-domain retrieval without per-task tuning. 
Controlled component ablations further validate the contribution of S$^2$LMR's core design choices (Appendix~\ref{app:core_ablation}). 
Together with the training-hyperparameter robustness of Section~\ref{sec:exp:stability:hp}, this shows that S$^2$LMR can be deployed on a new benchmark with minimal hyperparameter search.

\begin{figure}[t]
    \centering
    \includegraphics[width=\linewidth]{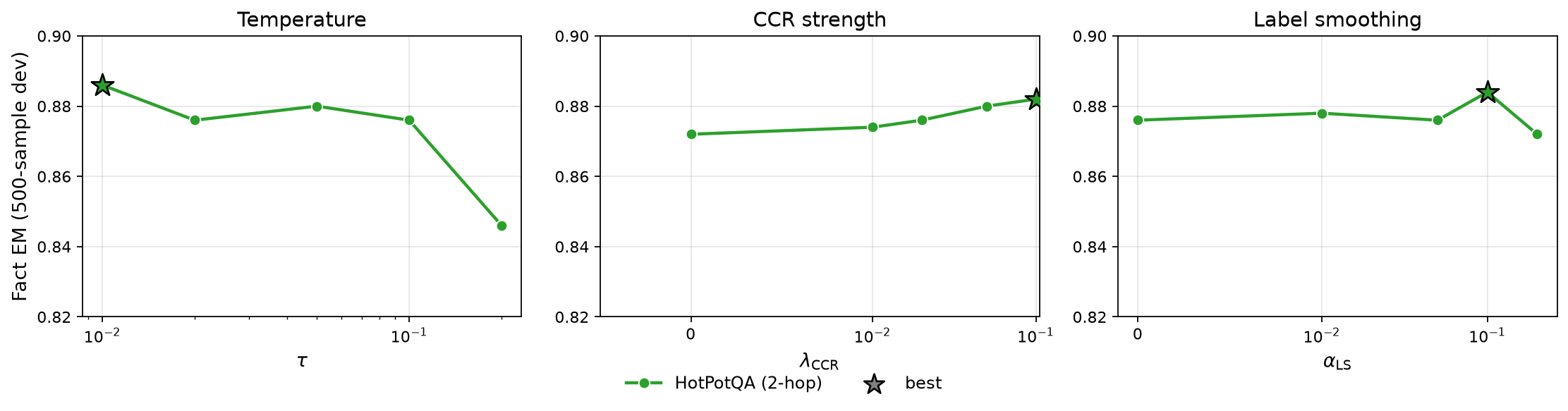}
    \caption{Fact EM vs (a) softmax temperature $\tau$ , (b) CCR strength $\lambda_{CCR}$, (c) label smoothing $\alpha_{LS}$. Stars denote per-task best.}
    \label{fig:ablation_hotpot}
\end{figure}

\begin{figure}
    \centering
    \includegraphics[width=1\linewidth]{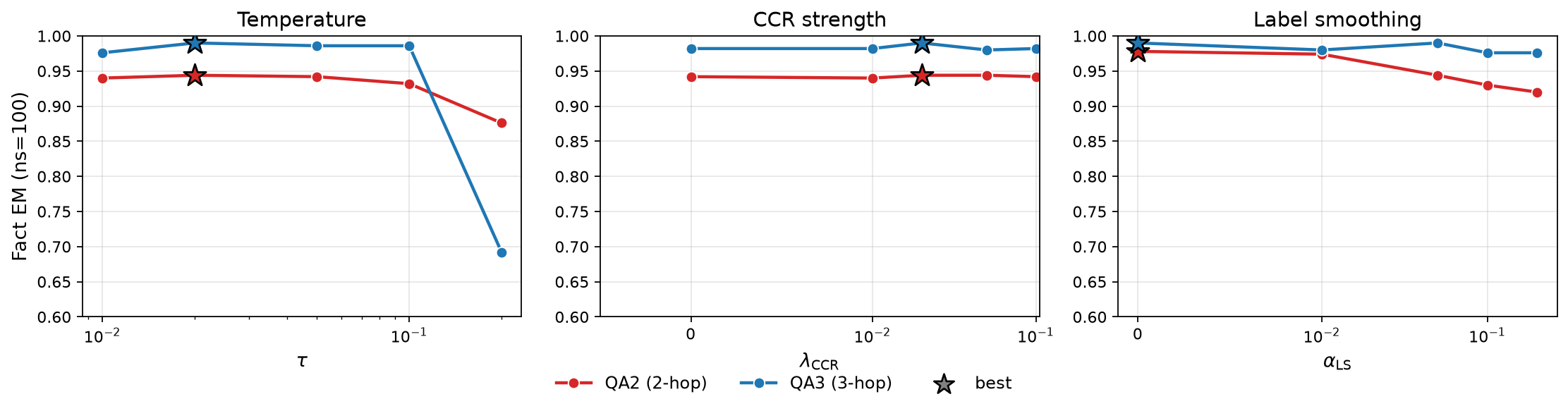}
    \caption{Fact EM vs (a) softmax temperature $\tau$, (b) CCR strength $\lambda_{CCR}$, (c) label smoothing $\alpha_{LS}$. Stars denote per-task best. S$^2$LMR maintains Fact EM $> 0.94$ (QA2) and $> 0.97$ (QA3) across a 10× range of each hyperparameter, degrading only at extreme values ($\tau = 0.2$ or $\alpha_{LS}\ge 0.05$).}
    \label{fig:ablation_babilong}
\end{figure}

\subsection{Leave-One-Out Ablation of Core Mechanisms}
\label{app:core_ablation}

Table~\ref{tab:core_ablation} reports Fact EM on BabiLong QA3 when each of the three components corresponding to Issue 1--Issue 3 identified in Section~\ref{sec:motivation_ssr} is individually removed from the full S$^2$LMR while all other design choices are held fixed.

\begin{table}[t]
    \centering
    \small
    \caption{Leave-one-out ablation of S$^2$LMR's three core mechanisms on BabiLong QA3 ($K{=}3$, strict Fact EM).}
    \label{tab:core_ablation}
    \begin{tabular}{l c c}
        \toprule
        Variant & Fact EM & $\Delta$EM \\
        \midrule
        S$^2$LMR (full)                                                                & \textbf{0.992} & --- \\
        \hspace{1em}$-$ Issue 1 (same-context $\to$ in-batch negatives)                & 0.28           & $-0.71$ \\
        \hspace{1em}$-$ Issue 2 (teacher-forced $\to$ beam-expanded per-hop, $B{=}2$)  & 0.948          & $-0.04$ \\
        \hspace{1em}$-$ Issue 3 (chunk-axis RoPE $\to$ no chunk-axis PE)               & 0.106          & $-0.89$ \\
        \bottomrule
    \end{tabular}
\end{table}

Removing same-context negatives (Issue 1) drops EM to $0.28$ ($\Delta\text{EM} = -0.71$), confirming that distractor-matched negative sampling is essential rather than incidental. 
Replacing teacher-forced per-hop supervision with beam-expanded per-hop supervision in the spirit of Beam Retriever~\citep{zhang2024end} (Issue 2) costs $0.04$ EM, showing that beam expansion of prev-hop contexts is neither necessary nor optimal. 
Removing chunk-axis RoPE (Issue 3) causes a near-total collapse of retrieval performance ($0.99 \to 0.11$, $\Delta\text{EM} = -0.89$), the largest single source of gain in this ablation.
Together, the three mechanisms are individually necessary and jointly account for essentially the entire gap between vanilla supervised bi-encoders and S$^2$LMR.

\subsection{Modern embedding backbone}
\label{app:qwen3}
\rev{We use the same backbones as Q-RAG for comparability. To test whether S$^2$LMR depends on the encoder,
we retrain it with Qwen3-Embedding-0.6B (last-token pooling, task instruction), using each benchmark's
original training budget and keeping all other hyperparameters fixed.}

\begin{table}[t]
\centering
\caption{\rev{Fact EM with the paper backbones and with Qwen3-Embedding-0.6B.}}
\label{tab:qwen3}
\begin{tabular}{@{}lcccc@{}}
\toprule
& \multicolumn{2}{c}{Paper backbone} & \multicolumn{2}{c}{Qwen3-Embedding-0.6B} \\
\cmidrule(lr){2-3}\cmidrule(lr){4-5}
& Zero-shot & S$^2$LMR & Zero-shot & S$^2$LMR \\
\midrule
BabiLong QA3 & 0.00 (Contriever) & 0.99 & 0.00 & 0.97 \\
HotpotQA     & 0.65 (e5-large)   & 0.90 & 0.54 & 0.90 \\
\bottomrule
\end{tabular}
\end{table}

\rev{S$^2$LMR transfers to the modern backbone without tuning, while a stronger off-the-shelf embedder alone
does not help: zero-shot Qwen3-Embedding is no better than the paper backbones.}

\rev{\paragraph{Issue~1 with a hard-negative-pretrained backbone.}
Hard negatives in modern embedders are mined offline per query for single-step retrieval. In multi-step
retrieval, however, a chunk's role depends on the hop: relevant facts, including later gold facts, act
as negatives until their turn. We repeat the Issue~1 ablation of Table~\ref{tab:core_ablation} with the Qwen3 backbone, with
chunk-axis PE in both arms. Same-context negatives reach 0.97 Fact EM on QA3, versus 0.36 with in-batch
negatives (Contriever, Table~\ref{tab:core_ablation}: 0.99 vs.\ 0.28). The gap persists, so the issue lies in the training
signal rather than in backbone strength.}

\section{Per-benchmark training hyperparameters}
\label{app:hparams}

Table~\ref{tab:hparams} lists the main training hyperparameters for
S$^2$LMR. Shared defaults across all runs:
per-sample per-hop InfoNCE with $\tau{=}0.02$, label smoothing
$\alpha_{\mathrm{LS}}{=}0.05$, and confidence-regularizer weight
$\lambda_{\mathrm{CCR}}{=}0.02$; AdamW with weight decay $0.01$ and
gradient clipping $1.0$; cosine LR schedule with $100$ warmup steps;
mean pooling with $\ell_2$-normalized embeddings; and greedy argmax
inference. 
At inference we set the retrieval horizon $T$ equal to $K$, the number of gold
supporting facts per query.
\rev{BabiLong and RULER NIAH runs use \texttt{facebook/contriever}, HotpotQA and MuSiQue use
\texttt{intfloat/multilingual-e5-large}. ``$\Delta$'' is the chunk-axis RoPE step (Eq.~2), ``--'' means
chunk-axis PE is disabled. Following Sorokin et al.~\citep{sorokin2026q}, the RULER SH/MH QA subtasks use a single
checkpoint trained on HotpotQA and MuSiQue with \texttt{Alibaba-NLP/gte-multilingual-base}, the backbone of
Q-RAG's released RULER QA checkpoint.
For Q-RAG, we evaluate the checkpoints released by \citet{sorokin2026q} on RULER QA, HotpotQA and MuSiQue, and retrain it with the configuration in the official repository on BabiLong, HotpotQA and MuSiQue. Where the published numbers were obtained under a different setting, we report both (Appendix~\ref{app:qrag_protocol}).} The stability comparison in Section~\ref{sec:exp:stability:hp}
additionally sweeps learning rate and effective batch size as described there.

\begin{table}[t]
\centering
\small
\caption{Per-benchmark training hyperparameters for S$^2$LMR.
$L_{\max}$: max tokens per chunk;
$\Delta$: chunk-axis RoPE step (Eq.~\ref{eq:pe}).}
\label{tab:hparams}
\begin{tabular}{lccccc}
\toprule
Benchmark & $L_{\max}$ & Batch & LR & Steps & $\Delta$ \\
\midrule
BabiLong QA1/2/4/5      & 256 & 8  & $1.5{\times}10^{-5}$  & 2000 & 20  \\
BabiLong QA3            & 64  & 32 & $1.06{\times}10^{-5}$ & 2000 & 200 \\
HotpotQA                & 256 & 8  & $1.5{\times}10^{-5}$  & 2000 & --  \\
MuSiQue                 & 512 & 8  & $1.5{\times}10^{-5}$  & 2500 & --  \\
RULER NIAH (S/MK/MV/MQ) & 256 & 8  & $1.5{\times}10^{-5}$  & 2000 & --  \\
\bottomrule
\end{tabular}
\end{table}

\textbf{Stability experiments.}
The training-stability comparison in Section~4.3 (Figures~5 and~6) uses
the same recipe as ``BabiLong QA3'' above with an extended schedule
matched to Q-RAG's reported setting: effective batch $64$
($32{\times}2$ grad accumulation), $\text{lr}{=}1.5{\times}10^{-5}$,
$50$K steps, linear LR schedule with $1000$ warmup steps, and weight
decay $0.1$.

\section{Fact-level metrics.}
\label{app:eval_details}
Let $S_{\mathrm{gt}}$ be the set of ground-truth supporting chunk indices
and $S_{\mathrm{pred}}$ the set of chunks returned by the retriever.
Following~\citep{sorokin2026q}, our Fact-EM metric is defined as
\[
    \mathrm{Fact\text{-}EM} =
    \begin{cases} 1, & \text{if } S_{\mathrm{gt}} \subseteq S_{\mathrm{pred}}, \\
                  0, & \text{otherwise}. \end{cases}
\]
Fact-EM gives full credit whenever the prediction covers all ground-truth chunks, even if it also contains additional, irrelevant ones. 
It does not require the two sets to be exactly equal.

\end{document}